\documentclass{article}
\newcommand{\papertitle}{Do Vision Transformers See Like Convolutional Neural Networks?}

\usepackage[square,numbers]{natbib}
\usepackage[final]{neurips_2021}

\author{%
  Maithra Raghu
  \\
  Google Research, Brain Team\\
  \texttt{maithrar@gmail.com} \\
  \And
  Thomas Unterthiner \\
  Google Research, Brain Team\\
  \texttt{unterthiner@google.com} \\
  \And
  Simon Kornblith \\
  Google Research, Brain Team\\
  \texttt{kornblith@google.com} \\
  \And
  Chiyuan Zhang \\
  Google Research, Brain Team\\
  \texttt{chiyuan@google.com} \\
  \And
  Alexey Dosovitskiy \\
  Google Research, Brain Team\\
  \texttt{adosovitskiy@google.com} \\
}

\usepackage[square,numbers]{natbib}

\usepackage[utf8]{inputenc} 
\usepackage[T1]{fontenc}    
\usepackage{hyperref}       
\usepackage{url}            
\usepackage{booktabs}       
\usepackage{amsfonts}       
\usepackage{nicefrac}       
\usepackage{microtype}      
\usepackage{xcolor}         

\usepackage{url}            
\usepackage{booktabs}       
\usepackage{amsfonts}       
\usepackage{nicefrac}       
\usepackage{microtype}      

\usepackage{graphicx}
\usepackage{url}
\usepackage{bm}
\usepackage[export]{adjustbox}
\usepackage{amsmath}
\usepackage{amsbsy}
\usepackage{calc}
\usepackage{amsthm}
\usepackage{amssymb}
\usepackage[percent]{overpic}
\usepackage[font=small,labelfont=bf]{caption}  
\usepackage{chngcntr}
\usepackage{placeins}
\usepackage{enumitem}

\usepackage{grffile}
\usepackage{subcaption}
\usepackage{booktabs} 
\usepackage{multirow}

\usepackage{sidecap}
\usepackage[export]{adjustbox}
\title{\papertitle}

\begin{document}

\maketitle

\begin{abstract}
  Convolutional neural networks (CNNs) have so far been the de-facto model for visual data. Recent work has shown that (Vision) Transformer models (ViT) can achieve comparable or even superior performance on image classification tasks. This raises a central question: \textit{how are Vision Transformers solving these tasks}? Are they acting like convolutional networks, or learning entirely different visual representations? Analyzing the internal representation structure of ViTs and CNNs on image classification benchmarks, we find striking differences between the two architectures, such as ViT having more uniform representations across all layers. We explore how these differences arise, finding crucial roles played by self-attention, which enables early aggregation of global information, and ViT residual connections, which strongly propagate features from lower to higher layers. We study the ramifications for spatial localization, demonstrating ViTs successfully preserve input spatial information, with noticeable effects from different classification methods. Finally, we study the effect of (pretraining) dataset scale on intermediate features and transfer learning, and conclude with a discussion on connections to new architectures such as the MLP-Mixer.    
\end{abstract}

\section{Introduction}
Over the past several years, the successes of deep learning on visual tasks has critically relied on convolutional neural networks \cite{krizhevsky2012imagenet, kolesnikov2019big}. This is largely due to the powerful inductive bias of spatial equivariance encoded by convolutional layers, which have been key to learning general purpose visual representations for easy transfer and strong performance. Remarkably however, recent work has demonstrated that Transformer neural networks are capable of equal or superior performance on image classification tasks at large scale \cite{dosovitskiy2020image}. These Vision Transformers (ViT) operate almost \textit{identically} to Transformers used in language \cite{devlin2018bert}, using self-attention, rather than convolution, to aggregate information across locations. 
This is in contrast with a large body of prior work, which has focused on more explicitly incorporating image-specific inductive biases \cite{parmar2018image, cordonnier2019relationship, carion2020end}

This breakthrough highlights a fundamental question: \textit{how} are Vision Transformers solving these image based tasks? Do they act like convolutions, learning the same inductive biases from scratch? Or are they developing novel task representations? What is the role of scale in learning these representations? And are there ramifications for downstream tasks? In this paper, we study these questions, uncovering key representational differences between ViTs and CNNs, the ways in which these difference arise, and effects on classification and transfer learning. Specifically, our contributions are: 
\begin{itemize}[leftmargin=1em,itemsep=0em,topsep=0em]
    \item We investigate the internal representation structure of ViTs and CNNs, finding striking differences between the two models, such as ViT having more uniform representations, with greater similarity between lower and higher layers.
    \item Analyzing how local/global spatial information is utilised, we find ViT incorporates more global information than ResNet at lower layers, leading to quantitatively different features.   
    \item Nevertheless, we find that incorporating local information at lower layers remains vital, with large-scale pre-training data helping early attention layers learn to do this 
    \item We study the uniform internal structure of ViT, finding that skip connections in ViT are even more influential than in ResNets, having strong effects on performance and representation similarity.
    \item Motivated by potential future uses in object detection, we examine how well input spatial information is preserved, finding connections between spatial localization and methods of classification.
    \item We study the effects of dataset scale on transfer learning, with a linear probes study revealing its importance for high quality intermediate representations.
\end{itemize}

\vspace{-0.25em}
\section{Related Work}
\vspace{-0.25em}
Developing non-convolutional neural networks to tackle computer vision tasks, particularly Transformer neural networks \cite{vaswani2017attention} has been an active area of research. Prior works have looked at \textit{local} multiheaded self-attention, drawing from the structure of convolutional receptive fields \cite{parmar2018image, ramachandran2019stand}, directly combining CNNs with self-attention \cite{carion2020end, bello2019attention, wu2020visual} or applying Transformers to smaller-size images \cite{chen2020generative, cordonnier2019relationship}. In comparison to these, the Vision Transformer \cite{dosovitskiy2020image} performs even less modification to the Transformer architecture, making it especially interesting to compare to CNNs. Since its development, there has also been very recent work analyzing aspects of ViT, particularly robustness \cite{bhojanapalli2021understanding, paul2021vision,naseer2021intriguing} and effects of self-supervision \cite{caron2021emerging,chen2021empirical}. Other recent related work has looked at designing hybrid ViT-CNN models \cite{yuan2021tokens, d2021convit}, drawing on structural differences between the models. Comparison between Transformers and CNNs are also recently studied in the text domain~\citep{tay2021pre}.

Our work focuses on the representational structure of ViTs. To study ViT representations, we draw on techniques from neural network representation similarity, which allow the quantitative comparisons of representations within and across neural networks \cite{kornblith2019similarity, raghu2017svcca, morcos2018insights, kriegeskorte2008representational}. These techniques have been very successful in providing insights on properties of different vision architectures \cite{nguyen2020wide, lindsay2020convolutional, kornblith2020s}, representation structure in language models \cite{wu2019emerging, merchant2020happens, wu2020similarity, kudugunta2019investigating}, dynamics of training methods \cite{raghu2019rapid, maheswaranathan2019universality} and domain specific model behavior \cite{mustafa2021supervised, raghu2019transfusion, shi2019comparison}. We also apply \emph{linear probes} in our study, which has been shown to be useful to analyze the learned representations in both vision~\citep{alain2016understanding} and text~\citep{conneau2018you,peters2018dissecting,voita2019bottom} models.

\vspace{-0.25em}
\section{Background and Experimental Setup}
\vspace{-0.25em}
\label{sec:background}
Our goal is to understand whether there are differences in the way ViTs represent and solve image tasks compared to CNNs. Based on the results of \citet{dosovitskiy2020image}, we take a representative set of CNN and ViT models --- ResNet50x1, ResNet152x2, ViT-B/32, ViT-B/16, ViT-L/16 and ViT-H/14. Unless otherwise specified, models are trained on the JFT-300M dataset \cite{sun2017revisiting}, although we also investigate models trained on the ImageNet ILSVRC 2012 dataset \cite{deng2009imagenet,russakovsky2015imagenet} and standard transfer learning benchmarks \cite{zhai2019visual, dosovitskiy2020image}.
We use a variety of analysis methods to study the layer representations of these models, gaining many insights into how these models function. 
We provide further details of the experimental setting in Appendix~\ref{app:experimental_setup}.

\textbf{Representation Similarity and CKA (Centered Kernel Alignment)}: Analyzing (hidden) layer representations of neural networks is challenging because their features are distributed across a large number of neurons. This distributed aspect also makes it difficult to meaningfully compare representations across neural networks. Centered kernel alignment (CKA)~\cite{kornblith2019similarity,cortes2012algorithms} addresses these challenges, enabling quantitative comparisons of representations within and across networks. Specifically, CKA takes as input $\mathbf{X} \in \mathbb{R}^{m \times p_1}$ and $\mathbf{Y} \in \mathbb{R}^{m \times p_2}$ which are representations (activation matrices), of two layers, with $p_1$ and $p_2$ neurons respectively, evaluated on the same $m$ examples. Letting $\bm{K} = \bm{X}\bm{X}^\mathsf{T}$ and $\bm{L} = \bm{Y}\bm{Y}^\mathsf{T}$ denote the Gram matrices for the two layers (which measures the similarity of a pair of datapoints according to layer representations) CKA computes:
\begin{align}
    \mathrm{CKA}(\bm{K}, \bm{L}) = \frac{\mathrm{HSIC}(\bm{K}, \bm{L})}{\sqrt{\mathrm{HSIC}(\bm{K}, \bm{K}) \mathrm{HSIC}(\bm{L}, \bm{L})}},
\end{align}
where $\mathrm{HSIC}$ is the Hilbert-Schmidt independence criterion~\cite{gretton2007kernel}. Given the centering matrix $\bm{H} = \bm{I}_n - \frac{1}{n}\bm{1}\bm{1}^\mathsf{T}$ and the centered Gram matrices  $\bm{K}' = \bm{H}\bm{K}\bm{H}$ and $\bm{L}' = \bm{H}\bm{L}\bm{H}$, $\mathrm{HSIC}(\bm{K},\bm{L}) = \mathrm{vec}(\bm{K}') \cdot \mathrm{vec}(\bm{L}')/(m-1)^2$, the similarity between these centered Gram matrices. CKA is invariant to orthogonal transformation of representations (including permutation of neurons), and the normalization term ensures invariance to isotropic scaling. These properties enable meaningful comparison and analysis of neural network hidden representations. To work at scale with our models and tasks, we approximate the unbiased estimator of $\mathrm{HSIC}$~\cite{song2012feature} using minibatches, as suggested in \cite{nguyen2020wide}.

\section{Representation Structure of ViTs and Convolutional Networks}
\label{sec:representation-structure}
\begin{figure}
    \centering
    \begin{tabular}{cccc}
    \hspace*{-10mm} \includegraphics[width=0.27\columnwidth]{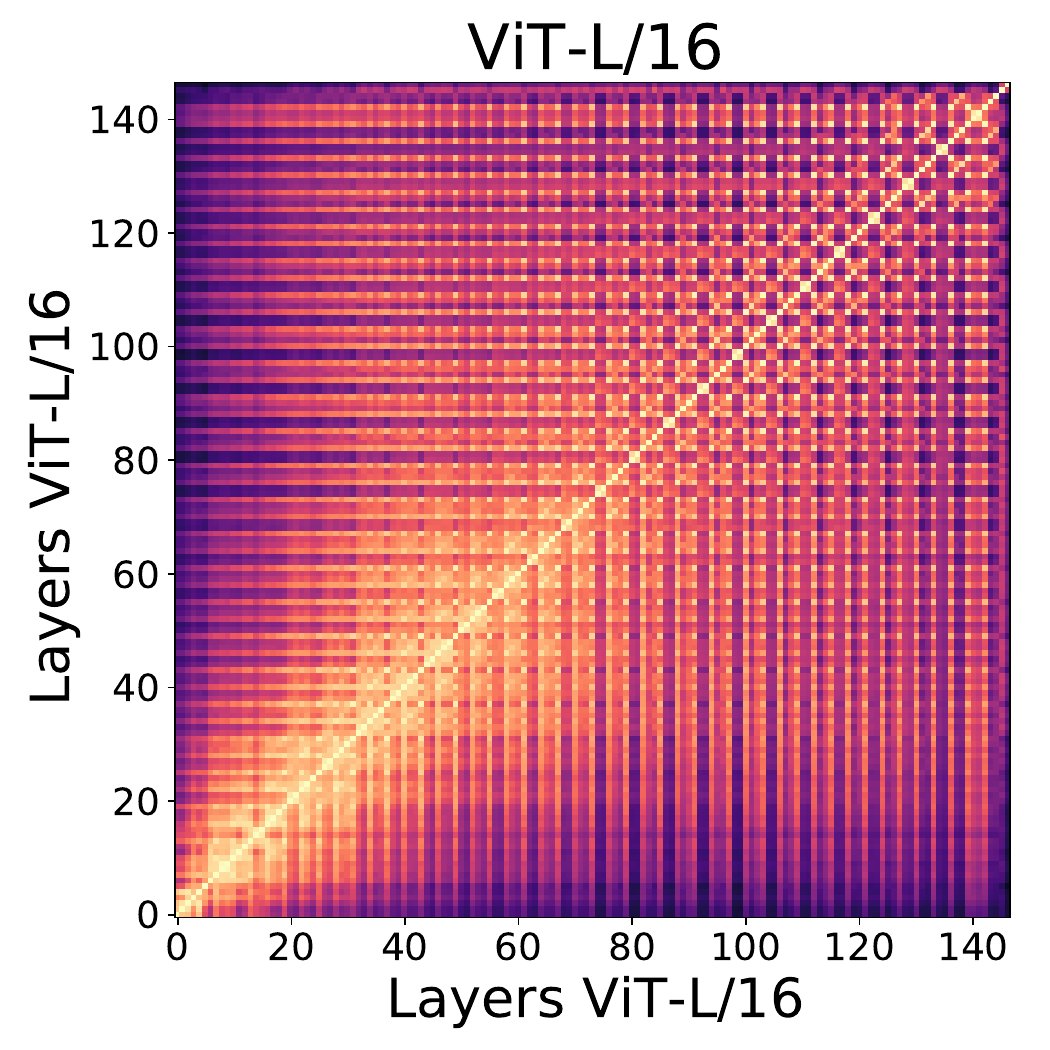} &
    \hspace*{-5mm} \includegraphics[width=0.27\columnwidth]{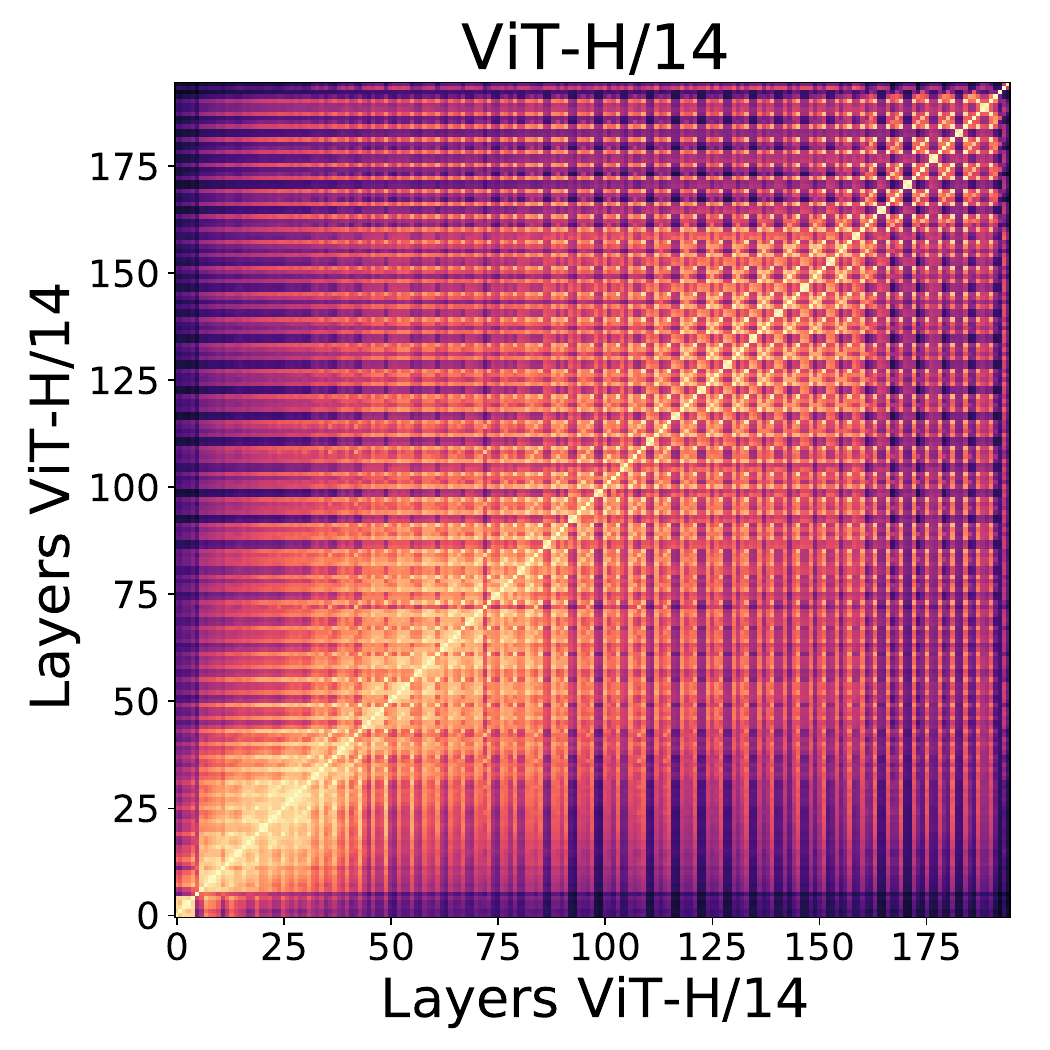} 
     \hspace{2mm}  \includegraphics[width=0.27\columnwidth]{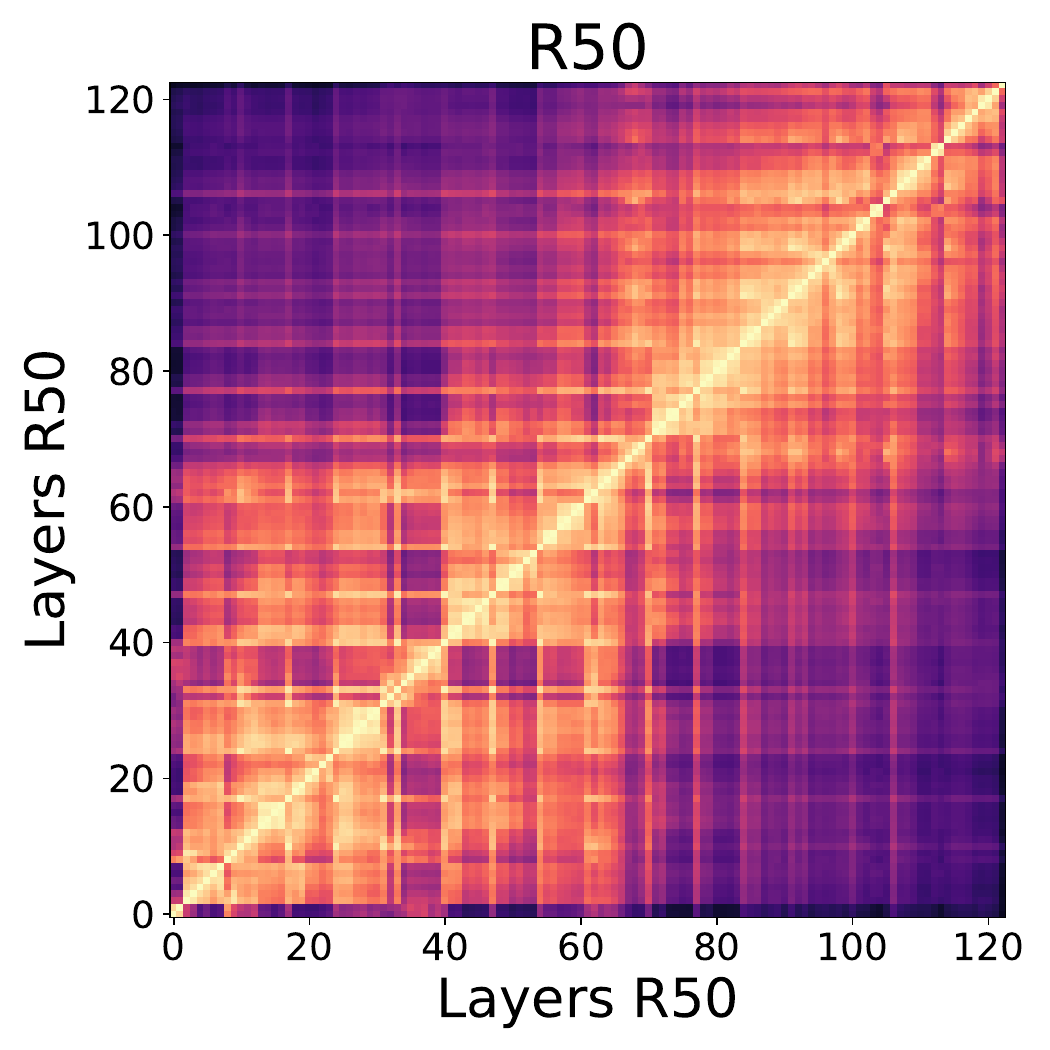} &
    \hspace*{-6mm} \includegraphics[width=0.27\columnwidth]{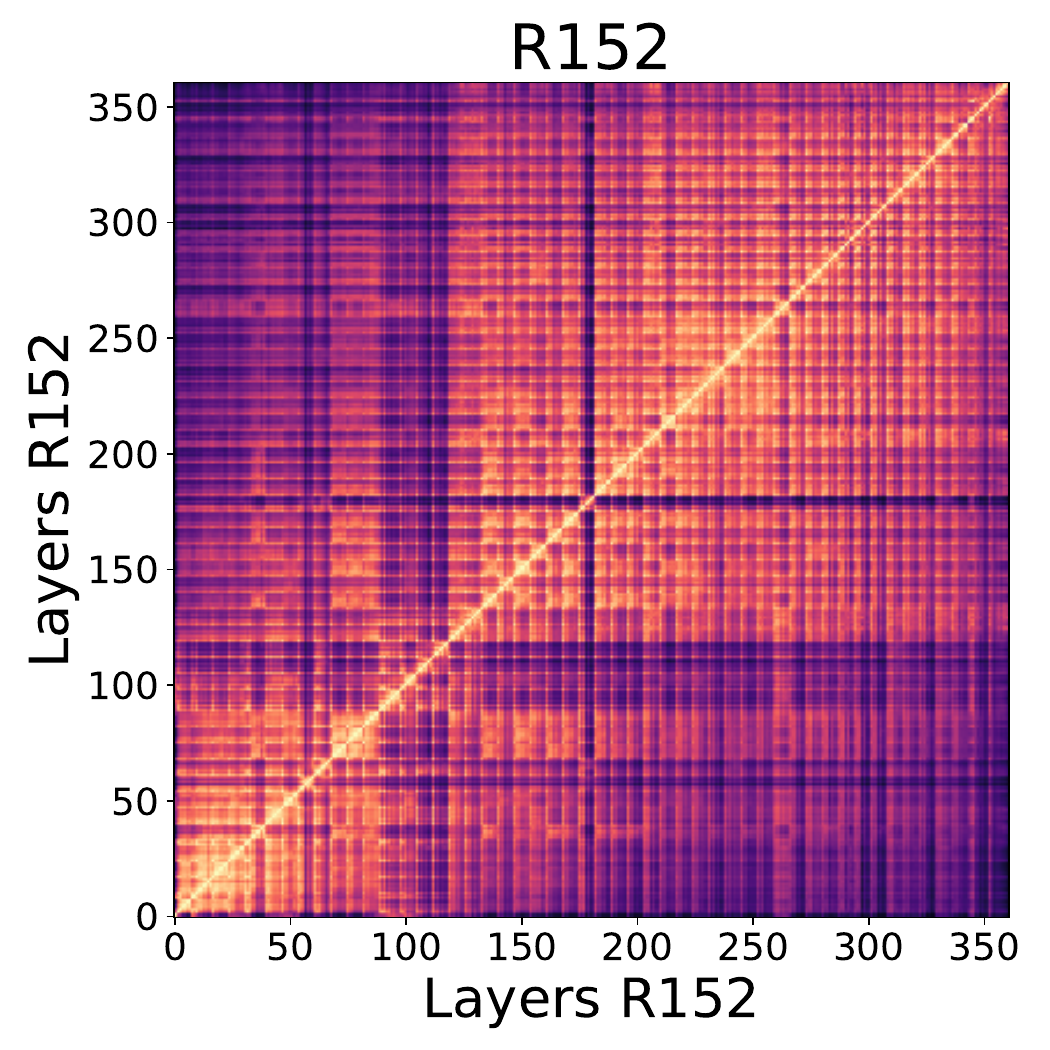}  \\
    \end{tabular}
    \caption{\small \textbf{Representation structure of ViTs and convolutional networks show significant differences, with ViTs having highly similar representations throughout the model, while the ResNet models show much lower similarity between lower and higher layers.} We plot CKA similarities between all pairs of layers across different model architectures. The results are shown as a heatmap, with the x and y axes indexing the layers from input to output. We observe that ViTs have relatively uniform layer similarity structure, with a clear grid-like pattern and large similarity between lower and higher layers. By contrast, the ResNet models show clear stages in similarity structure, with smaller similarity scores between lower and higher layers.}
    \label{fig:representation-structure}
\end{figure}
We begin our investigation by using CKA to study the internal representation structure of each model. How are representations propagated within the two architectures, and are there signs of functional differences? 
To answer these questions, we take every pair of layers $\bm{X}, \bm{Y}$ within a model and compute their CKA similarity. Note that we take representations not only from outputs of ViT/ResNet blocks, but also from intermediate layers, such as normalization layers and the hidden activations inside a ViT MLP. Figure \ref{fig:representation-structure} shows the results as a heatmap, for multiple ViTs and ResNets. We observe clear differences between the internal representation structure between the two model architectures: (1) ViTs show a much more uniform similarity structure, with a clear grid like structure (2) lower and higher layers in ViT show much greater similarity than in the ResNet, where similarity is divided into different (lower/higher) stages.

\begin{SCfigure}[50]
    \centering
    \begin{tabular}{cc}
     \hspace{-16pt}\includegraphics[width=0.35\linewidth]{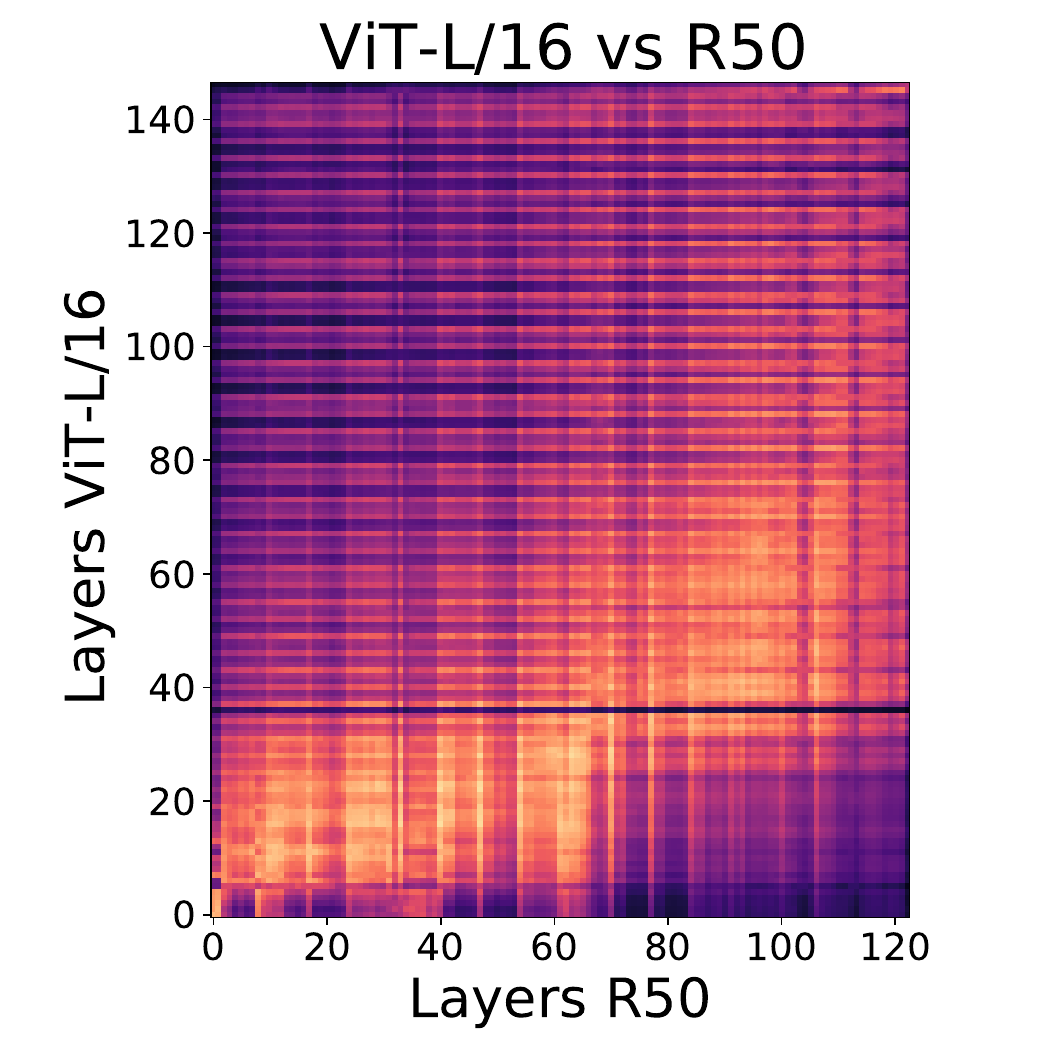} &
     \hspace{-32pt}\includegraphics[width=0.35\linewidth]{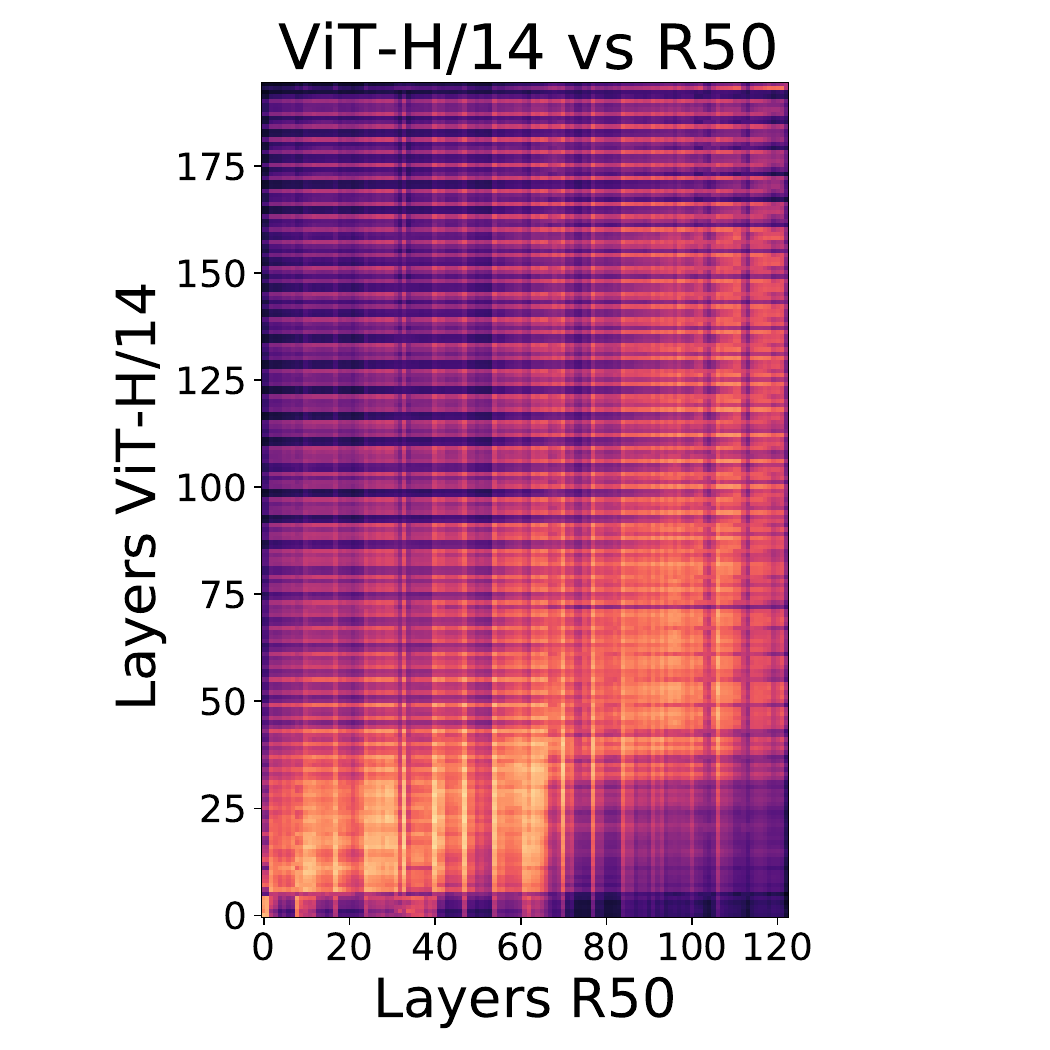} \\
    \end{tabular}\hspace{-31pt}\vspace{-6pt}
    \caption{\textbf{Cross model CKA heatmap between ViT and ResNet illustrate that a larger number of lower layers in the ResNet are similar to a smaller set of the lowest ViT layers.} We compute a CKA heatmap comparing all layers of ViT to all layers of ResNet, for two different ViT models. We observe that the lower half of ResNet layers are similar to around the lowest quarter of ViT layers. The remaining half of the ResNet is similar to approximately the next third of ViT layers, with the highest ViT layers dissimilar to lower and higher ResNet layers.}
    \label{fig:representation-cross-comparisons}
\end{SCfigure}

We also perform cross-model comparisons, where we take all layers $\bm{X}$ from ViT and compare to all layers $\bm{Y}$ from ResNet. We observe (Figure \ref{fig:representation-cross-comparisons}) that the lower half of ~60 ResNet layers are similar to approximately the lowest quarter of ViT layers. In particular, many more lower layers in the ResNet are needed to compute similar representations to the lower layers of ViT. The top half of the ResNet is approximately similar to the next third of the ViT layers. The final third of ViT layers is less similar to all ResNet layers, likely because this set of layers mainly manipulates the CLS token representation, further studied in Section \ref{sec:skip-connections}.

Taken together, these results suggest that (i) ViT lower layers compute representations in a different way to lower layers in the ResNet, (ii) ViT also more strongly propagates representations between lower and higher layers (iii) the highest layers of ViT have quite different representations to ResNet.  

\section{Local and Global Information in Layer Representations}
\label{sec:local-global}
In the previous section, we observed much greater similarity between lower and higher layers in ViT, and we also saw that ResNet required more lower layers to compute similar representations to a smaller set of ViT lower layers. In this section, we explore one possible reason for this difference: the difference in the ability to incorporate global information between the two models. How much global information is aggregated by early self-attention layers in ViT? Are there noticeable resulting differences to the features of CNNs, which have fixed, local receptive fields in early layers? In studying these questions, we demonstrate the influence of global representations and a surprising connection between scale and self-attention distances.

\textbf{Analyzing Attention Distances:}
\begin{figure}
\centering
    \begin{tabular}{cc}
     \includegraphics[width=0.4\columnwidth]{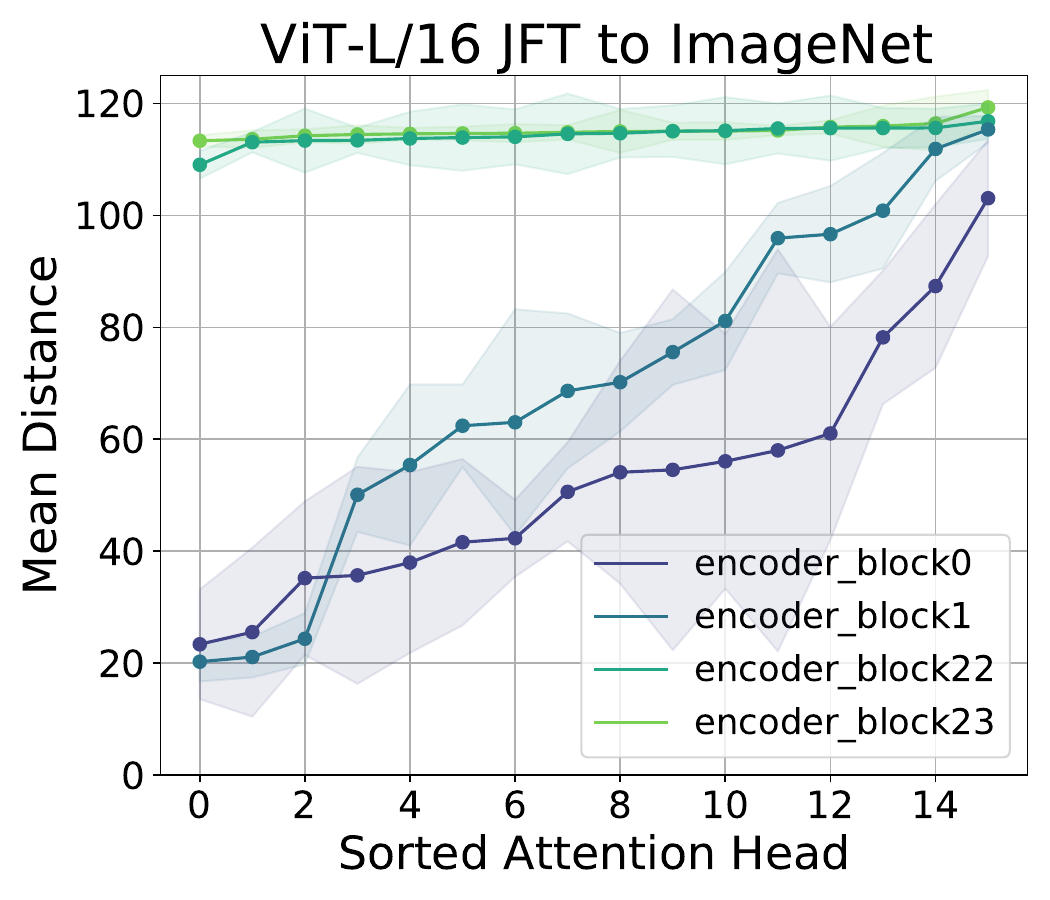} &
     \includegraphics[width=0.4\columnwidth]{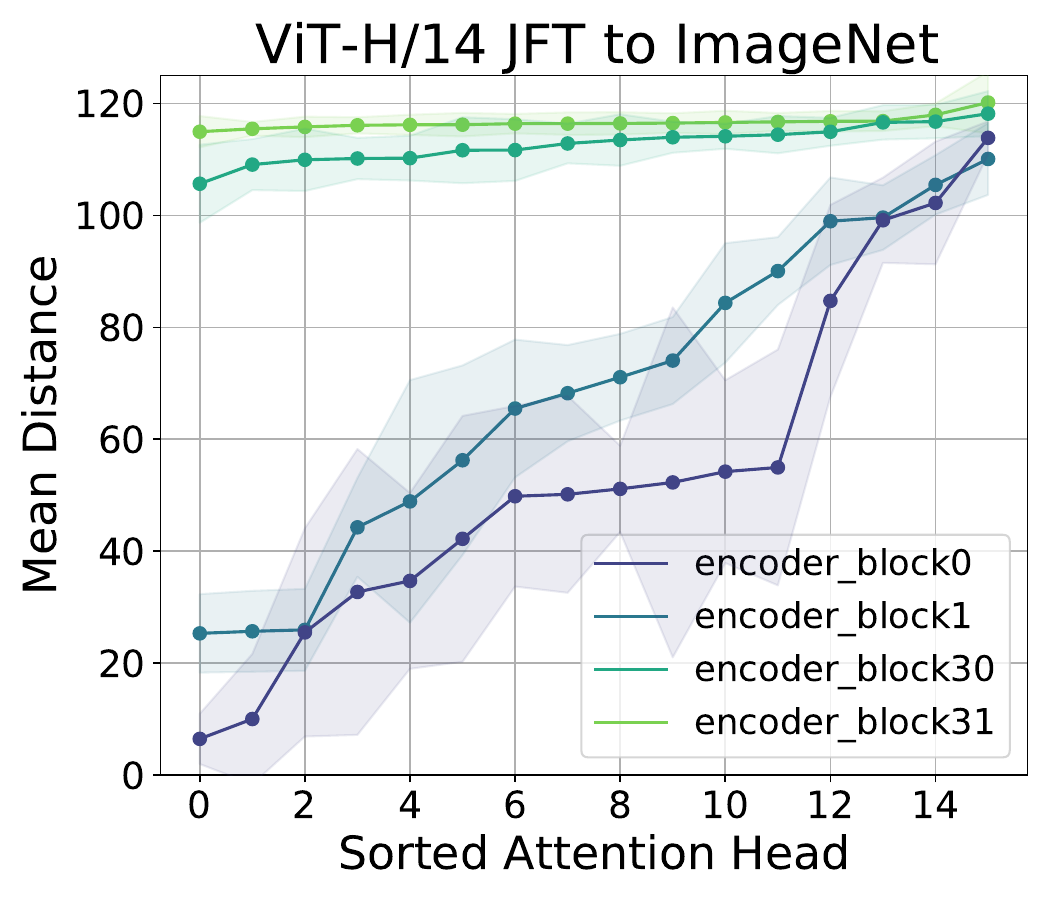} \\
    \end{tabular}
    \caption{\small \textbf{Plotting attention head mean distances shows lower ViT layers attend both locally and globally, while higher layers primarily incorporate global information.} For each attention head, we compute the pixel distance it attends to, weighted by the attention weights, and then average over 5000 datapoints to get an average attention head distance. We plot the heads sorted by their average attention distance for the two lowest and two highest layers in the ViT, observing that the lower layers attend both locally and globally, while the higher layers attend entirely globally. }
    \label{fig:attention-heads-plot}
\end{figure}
We start by analyzing ViT self-attention layers, which are the mechanism for ViT to aggregate information from other spatial locations, and structurally very different to the fixed receptive field sizes of CNNs. Each self-attention layer comprises multiple self-attention heads, and for each head we can compute the average distance between the query patch position and the locations it attends to.
This reveals how much local vs global information each self-attention layer is aggregating for the representation.
Specifically, we weight the pixel distances by the attention weights for each attention head and average over 5000 datapoints, with results shown in Figure \ref{fig:attention-heads-plot}. 
In agreement with~\citet{dosovitskiy2020image}, we observe that even in the lowest layers of ViT, self-attention layers have a mix of local heads (small distances) and global heads (large distances).
This is in contrast to CNNs, which are hardcoded to attend only locally in the lower layers.
At higher layers, all self-attention heads are global.

\begin{figure}
\centering
    \begin{tabular}{cc}
    \includegraphics[width=0.4\columnwidth]{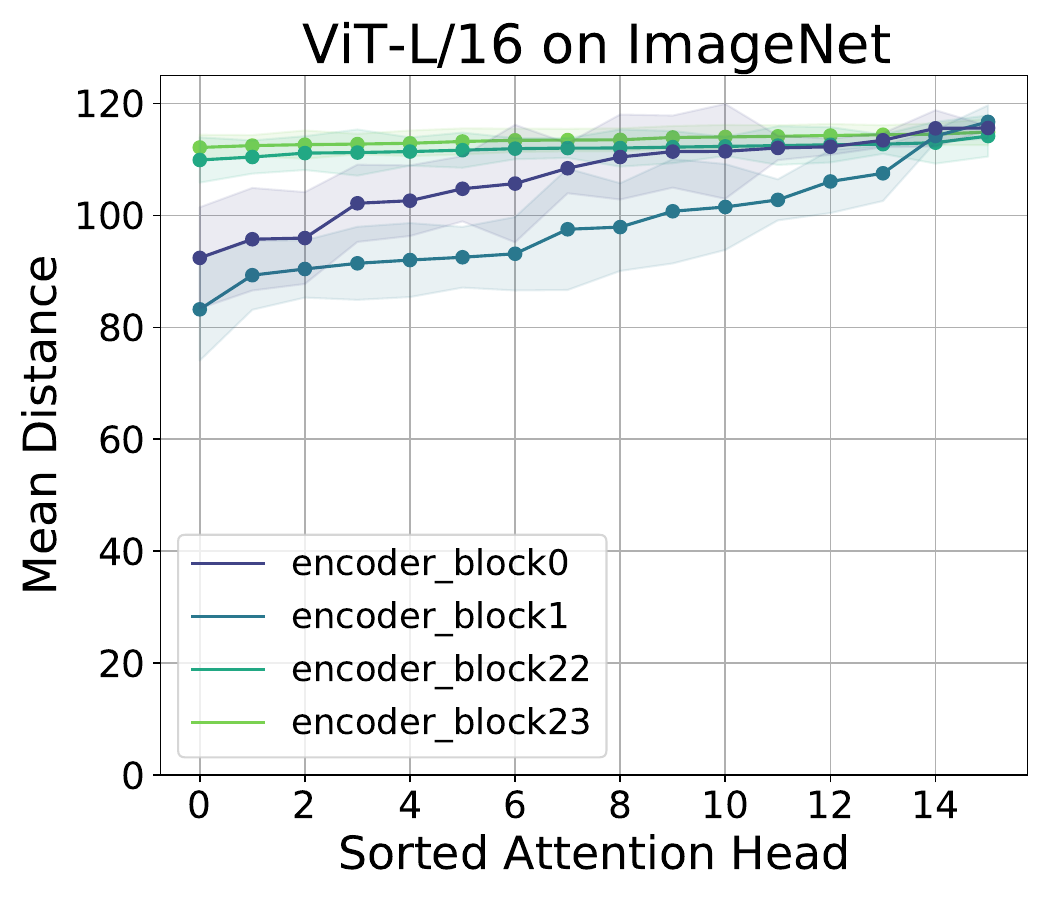} &
    \includegraphics[width=0.4\columnwidth]{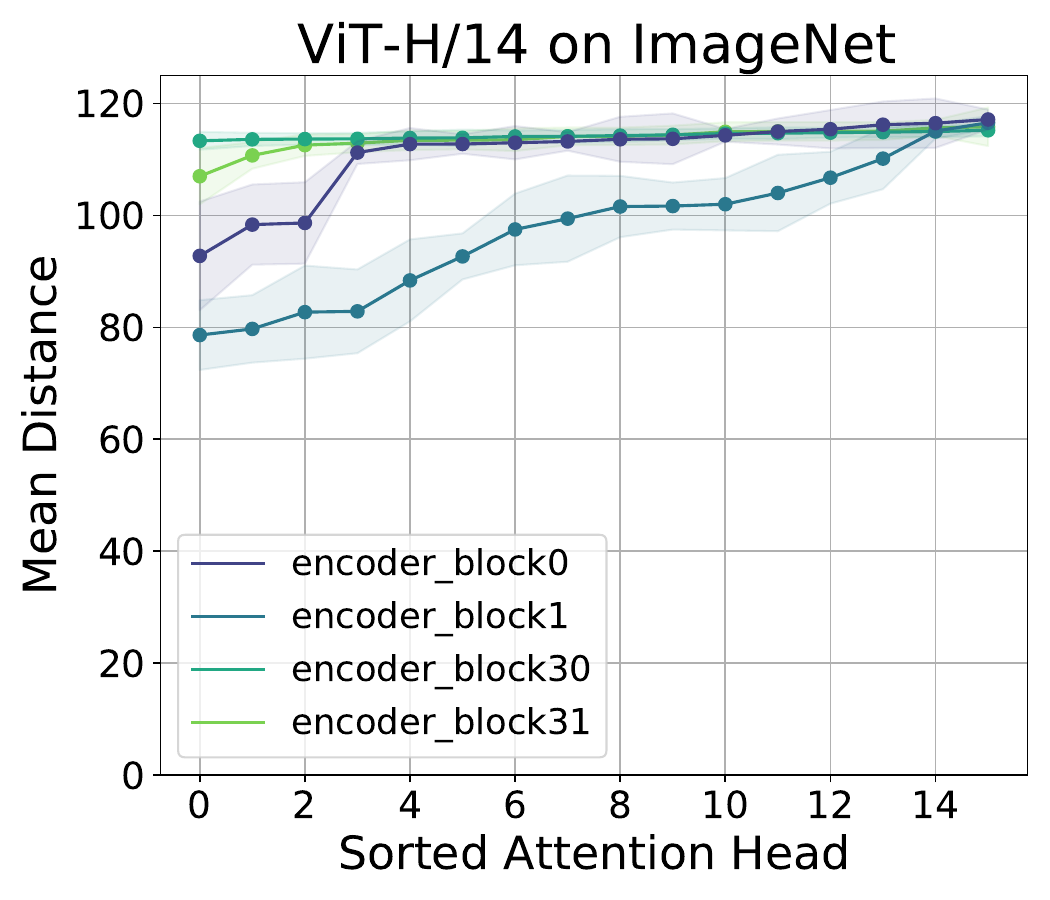} \\
    \end{tabular}
    \caption{\small \textbf{With less training data, lower attention layers do not learn to attend locally.} Comparing the results to Figure \ref{fig:attention-heads-plot}, we see that training only on ImageNet leads to the lower layers not learning to attend more locally. These models also perform much worse when only trained on ImageNet, suggesting that incorporating local features (which is hardcoded into CNNs) may be important for strong performance. (See also Figure \ref{fig:attention-heads-ViTB-app}.)}
    \label{fig:attention-heads-scale}
\end{figure}

Interestingly, we see a clear effect of scale on attention. In Figure \ref{fig:attention-heads-scale}, we look at attention distances when training \textit{only} on ImageNet (no large-scale pre-training), which leads to much lower performance in ViT-L/16 and ViT-H/14 \cite{dosovitskiy2020image}. Comparing to Figure \ref{fig:attention-heads-plot}, we see that with not enough data, ViT \textit{does not learn to attend locally} in earlier layers. Together, this suggests that using local information early on for image tasks (which is hardcoded into CNN architectures) is important for strong performance.

\textbf{Does access to global information result in different features?}
The results of Figure \ref{fig:attention-heads-plot} demonstrate that ViTs have access to more global information than CNNs in their lower layers. But does this result in different learned features? As an interventional test, we take subsets of the ViT attention heads from the first encoder block, ranging from the subset corresponding to the most local attention heads to a subset of the representation corresponding to the most global attention heads. We then compute CKA similarity between these subsets and the lower layer representations of ResNet.

\begin{figure}[t!]
\centering
    \begin{tabular}{ccc}
    \hspace*{-7mm} \includegraphics[width=0.34\columnwidth]{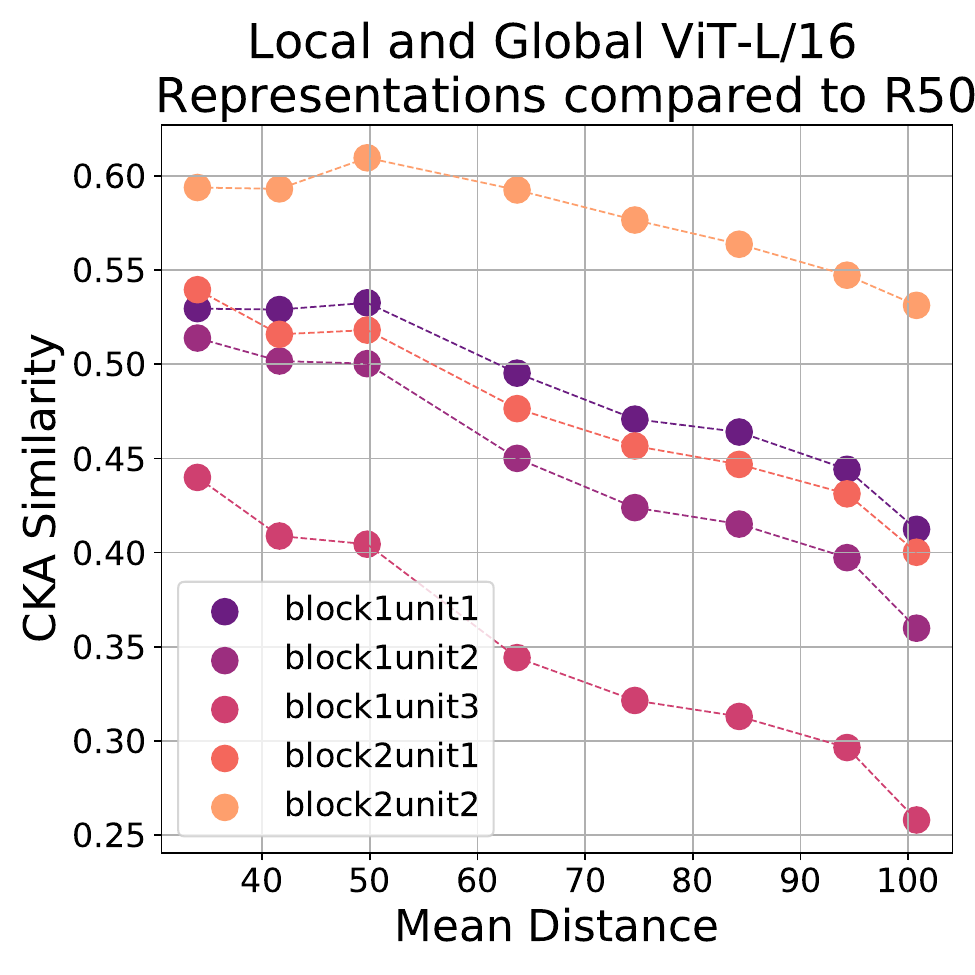} &
    \hspace*{-5mm} \includegraphics[width=0.34\columnwidth]{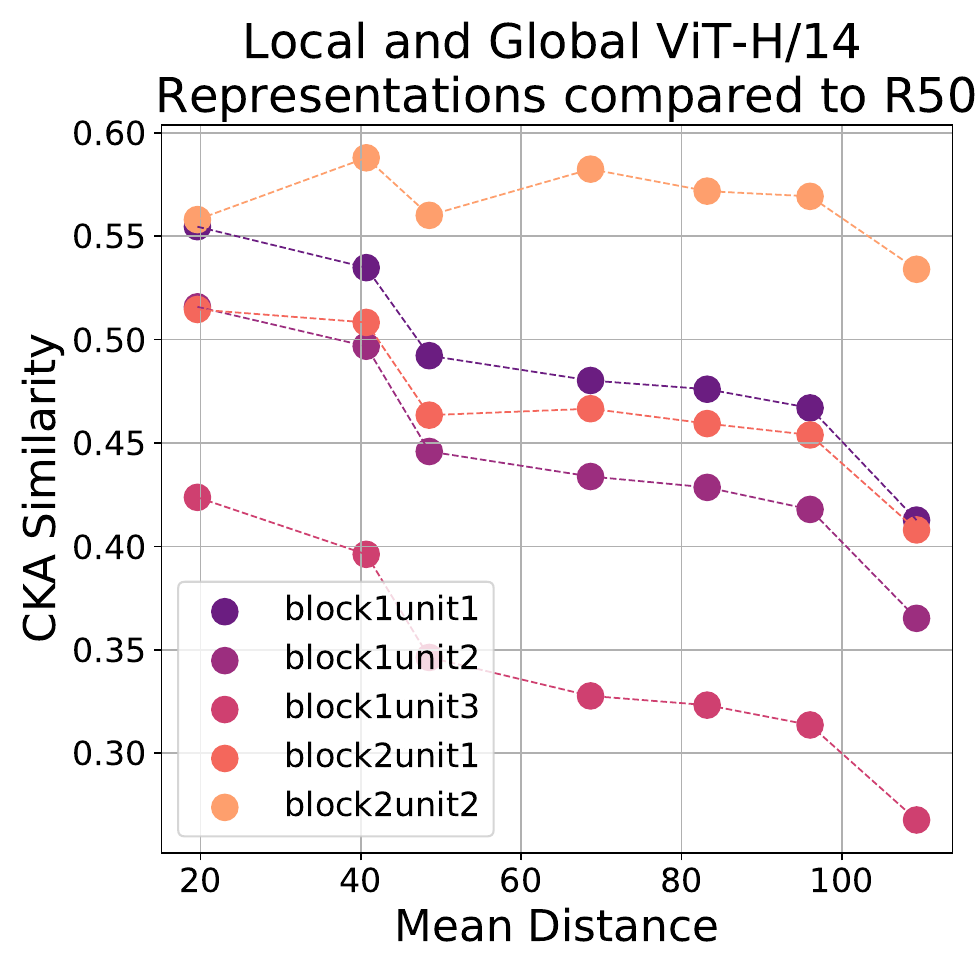} &
    \hspace*{-5mm} \includegraphics[width=0.34\columnwidth]{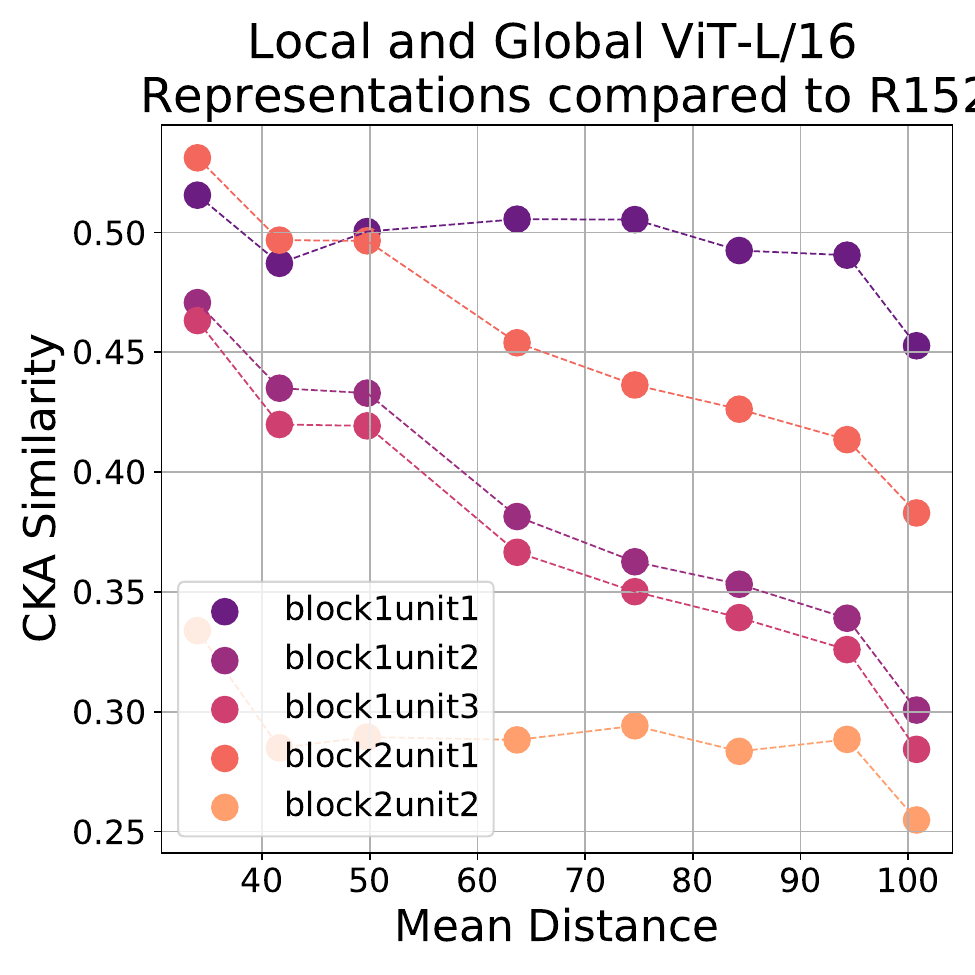} \\
    \end{tabular}
    \caption{\small \textbf{Lower layer representations of ResNet are most similar to representations corresponding to local attention heads of ViT.} We take subsets of ViT attention heads in the first encoder block, ranging from the most locally attending heads (smallest mean distance) to the most global heads (largest mean distance). We then compute CKA similarity between these subsets and lower layer representations in the ResNet. We observe that lower ResNet layers are most similar to the features learned by local attention heads of ViT, and decrease monotonically in similarity as more global information is incorporated, demonstrating that the global heads learn quantitatively different features.}
    \label{fig:local-global-split}
\end{figure}
The results, shown in Figure \ref{fig:local-global-split}, which plot the mean distance for each subset against CKA similarity, clearly show a monotonic decrease in similarity as mean attention distance grows, demonstrating that access to more global information also leads to quantitatively different features than computed by the local receptive fields in the lower layers of the ResNet. 

\begin{SCfigure}[50]
    \centering
    \hspace{-1.5em}
    \begin{tabular}{c}
    \includegraphics[width=0.7\linewidth]{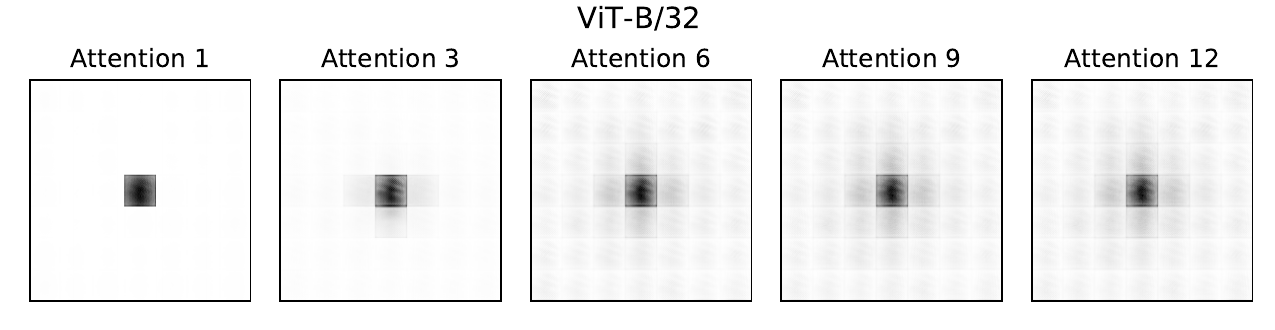}\\
    \includegraphics[width=0.7\linewidth]{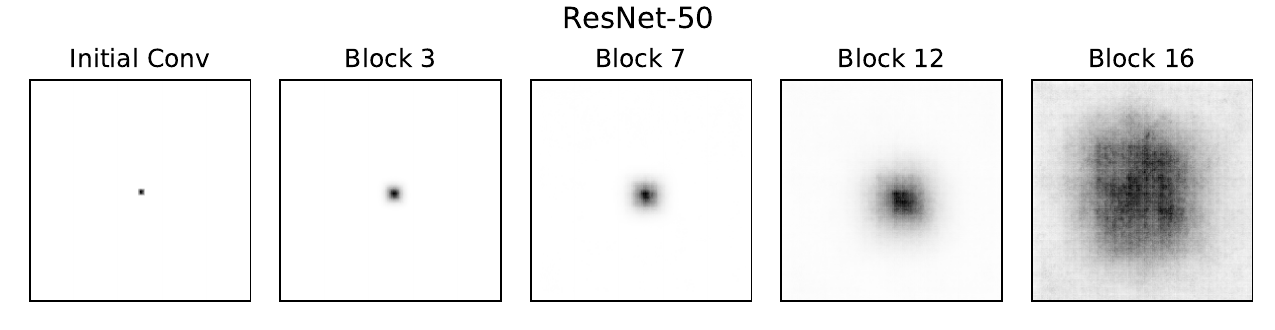}
    \end{tabular}
    \hspace{-1.0em}
    \caption{\small \textbf{ResNet effective receptive fields are highly local and grow gradually; ViT effective receptive fields shift from local to global}. We measure the effective receptive field of different layers as the absolute value of the gradient of the center location of the feature map (taken after residual connections) with respect to the input. 
    Results are averaged across all channels in each map for 32 randomly-selected images.
    }
    \label{fig:receptive-fields-summary}
\end{SCfigure}

\textbf{Effective Receptive Fields:}
We conclude by computing \textit{effective receptive fields} \cite{luo2017understanding} for both ResNets and ViTs, with results in Figure \ref{fig:receptive-fields-summary} and Appendix \ref{app:local-global}. We observe that lower layer effective receptive fields for ViT are indeed larger than in ResNets, and while ResNet effective receptive fields grow gradually, ViT receptive fields become much more global midway through the network. ViT receptive fields also show strong dependence on their center patch due to their strong residual connections, studied in the next section. As we show in Appendix~\ref{app:local-global}, in attention sublayers, receptive fields taken before the residual connection show far less dependence on this central patch.

\section{Representation Propagation through Skip Connections}
\label{sec:skip-connections}
The results of the previous section demonstrate that ViTs learn different representations to ResNets in lower layers due to access to global information, which explains some of the differences in representation structure observed in Section \ref{sec:representation-structure}. However, the highly uniform nature of ViT representations (Figure \ref{fig:representation-structure}) also suggests lower representations are faithfully propagated to higher layers. But how does this happen? In this section, we explore the role of skip connections in representation propagation across ViTs and ResNets, discovering ViT skip connections are highly influential, with a clear phase transition from preserving the CLS (class) token representation (in lower layers) to spatial token representations (in higher layers).

\begin{figure}[t!]
    \centering
    \begin{tabular}{cc}
        \includegraphics[width=0.45\columnwidth]{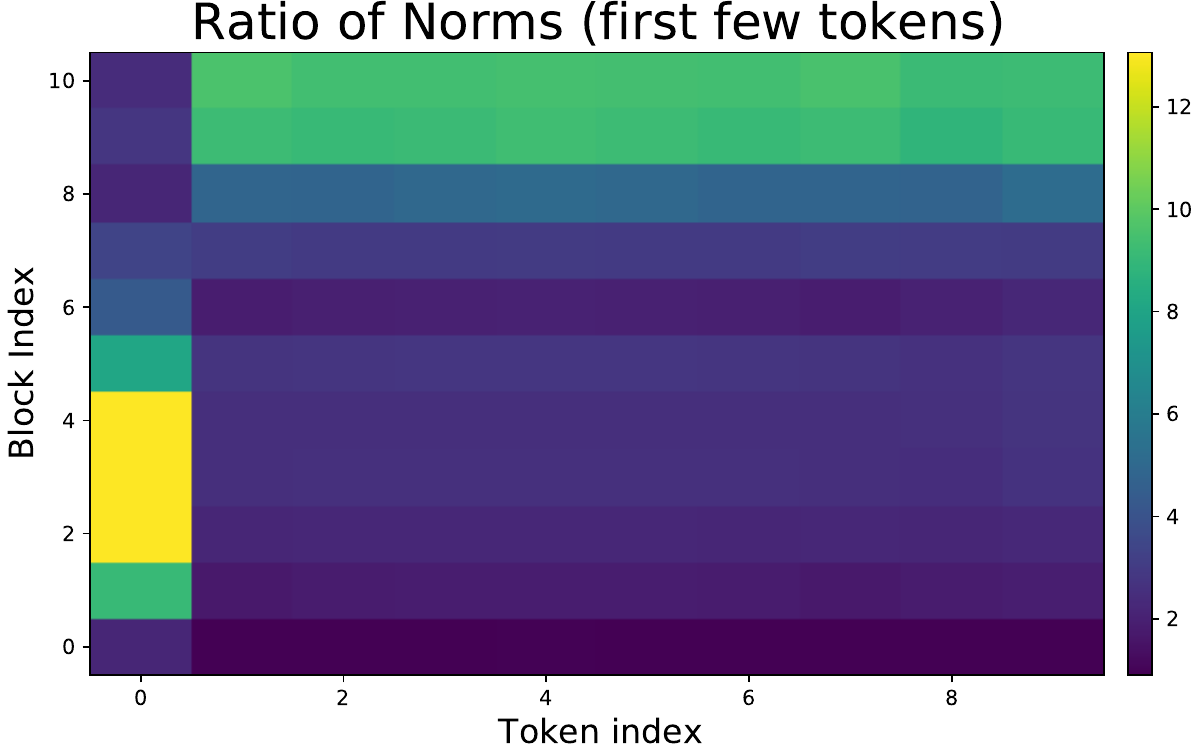} ~
        \includegraphics[width=0.45\columnwidth]{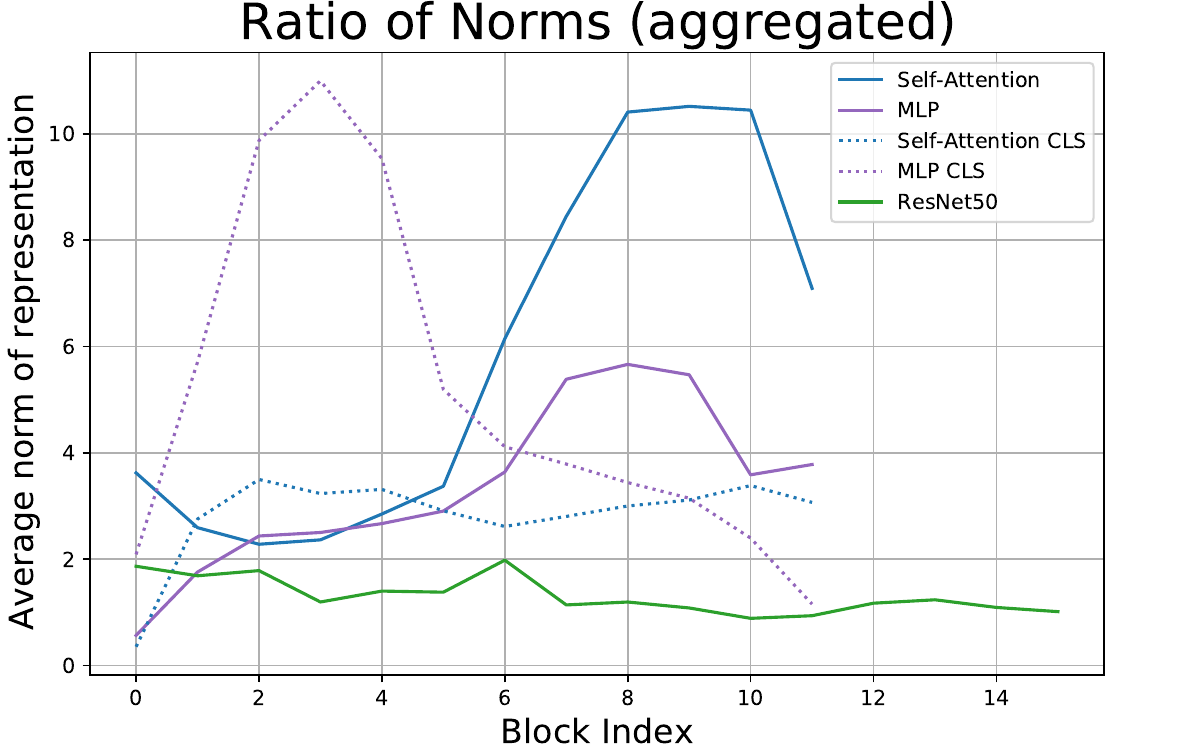}
    \end{tabular}
    \caption{\textbf{Most information in ViT passes through skip connections}. Comparison of representation norms between the skip-connection (identity) and the long branch for ViT-B/16 trained on ImageNet and a ResNet. For ViT, we show the CLS token separately from the rest of the representation. (left) shows the ratios separated for the first few tokens (token 0 is CLS), (right) shows averages over all tokens.}
    \label{fig:identity-connection-norms-vit}

    \label{fig:identity-connection-norms-line-plot}
\end{figure}

Like Transformers, ViTs contain \textit{skip} (aka \textit{identity} or \textit{shortcut}) connections throughout, which are added on after the (i) self-attention layer, and (ii) MLP layer. To study their effect, we plot the norm ratio $||z_i||/||f(z_i)||$ where $z_i$ is the hidden representation of the $i$th layer coming from the skip connection, and $f(z_i)$ is the transformation of $z_i$ from the \textit{long branch} (i.e.  MLP or self-attention.) 

The results are in Figure \ref{fig:identity-connection-norms-vit} (with additional cosine similarity analysis in Figure \ref{fig:identity-connection-cosine-similarity-plot}.) The heatmap on the left shows $||z_i||/||f(z_i)||$ for different token representations. We observe a striking phase transition: in the first half of the network, the CLS token (token 0) representation is primarily propagated by the skip connection branch (high norm ratio), while the spatial token representations have a large contribution coming from the long branch (lower norm ratio). Strikingly, in the second half of the network, this is reversed. 

The right pane, which has line plots of these norm ratios across ResNet50, the ViT CLS token and the ViT spatial tokens additionally demonstrates that skip connection is much more influential in ViT compared to ResNet: we observe much higher norm ratios for ViT throughout, along with the phase transition from CLS to spatial token propagation (shown for the MLP and self-attention layers.)

\textbf{ViT Representation Structure without Skip Connections:}
\begin{figure}
    \centering
     \includegraphics[width=0.8\columnwidth]{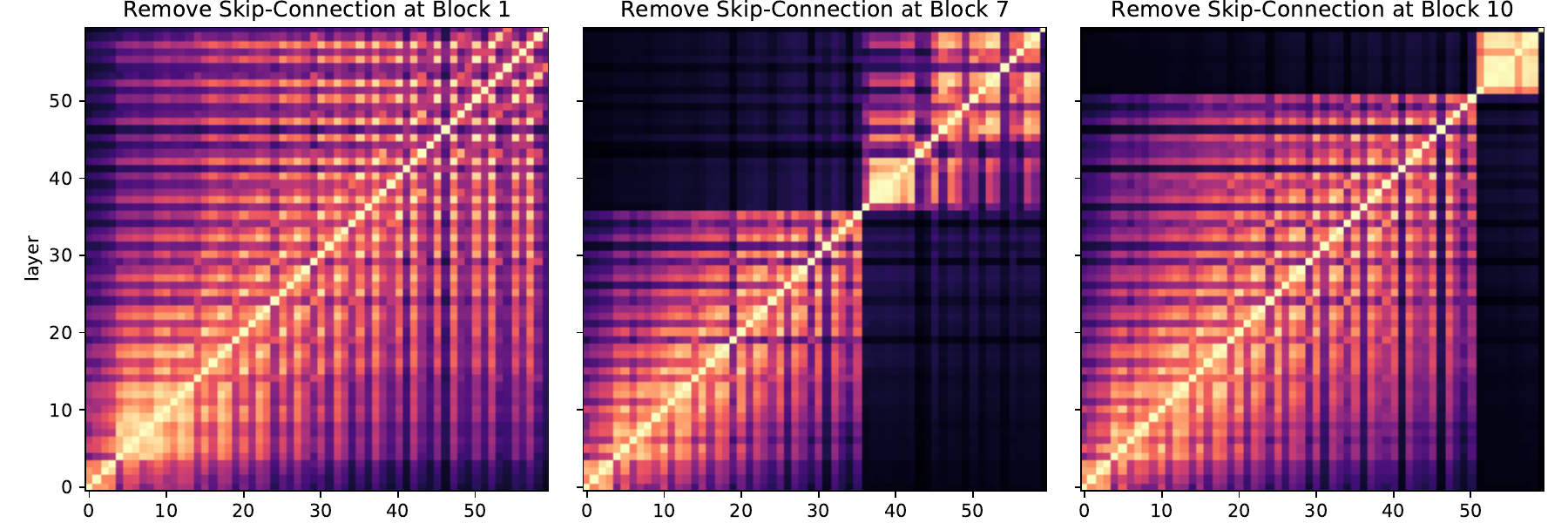}
    \caption{\small \textbf{ViT models trained without any skip connections in block $i$ show very little representation similarity between layers before/after block $i$.} We train several ViT models without any skip connections at block $i$ for varying $i$ to interventionally test the effect on representation structure. For middle blocks without skip connections, we observe a performance drop of ~$4\%$. We also observe that removing a skip connection at block $i$ partitions similar representations to before/after block $i$ --- this demonstrates the importance of skip connections in ViT's standard uniform representation structure.}
    \label{fig:identity-connection-norms-cka}
    \vspace{-1.5em}
\end{figure}
The norm ratio results strongly suggest that skip connections play a key role in the representational structure of ViT. To test this interventionally, we train ViT models with skip connections removed in block $i$ for varying $i$, and plot the CKA representation heatmap. The results, in Figure \ref{fig:identity-connection-norms-cka}, illustrate that removing the skip connections in a block partitions the layer representations on either side. (We note a performance drop of ~$4\%$ when removing skip connections from middle blocks.) This demonstrates the importance of representations being propagated by skip connections for the uniform similarity structure of ViT in Figure \ref{fig:representation-structure}.

\section{Spatial Information and Localization}
\label{sec:spatial-localization}
The results so far, on the role of self-attention in aggregating spatial information in ViTs, and skip-connections faithfully propagating representations to higher layers, suggest an important followup question: how well can ViTs perform \textit{spatial localization}? Specifically, is spatial information from the input preserved in the higher layers of ViT? And how does it compare in this aspect to ResNet? An affirmative answer to this is crucial for uses of ViT beyond classification, such as object detection.

\begin{figure}[t]
    \centering
    \includegraphics[width=0.8\columnwidth]{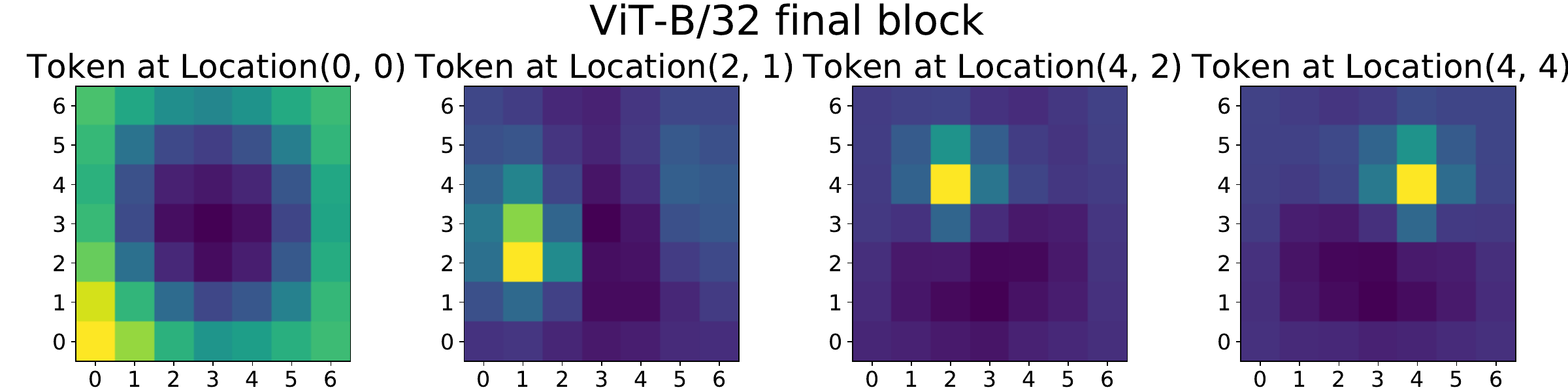} 
    \includegraphics[width=0.8\columnwidth]{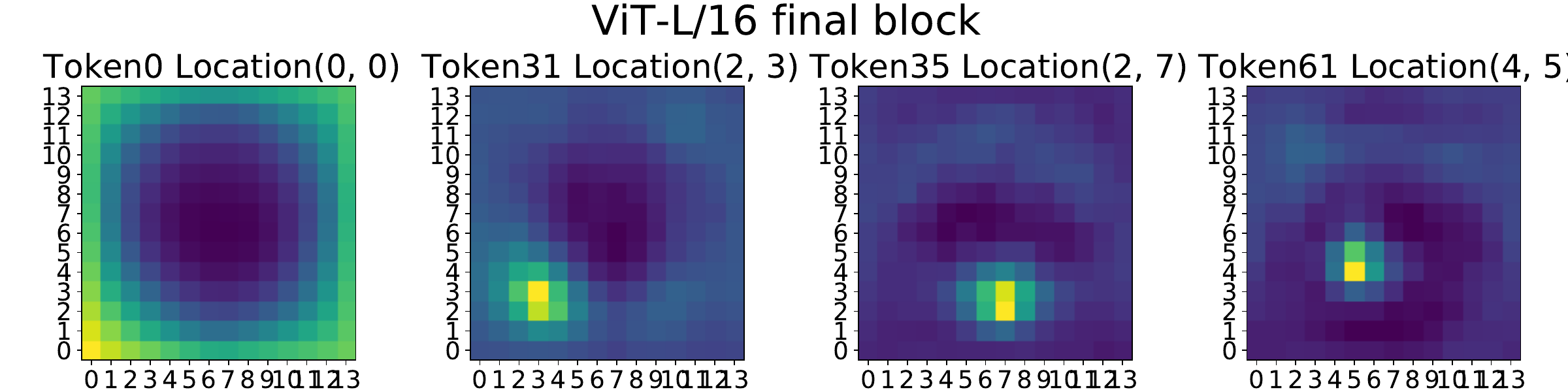}
    \includegraphics[width=0.8\columnwidth]{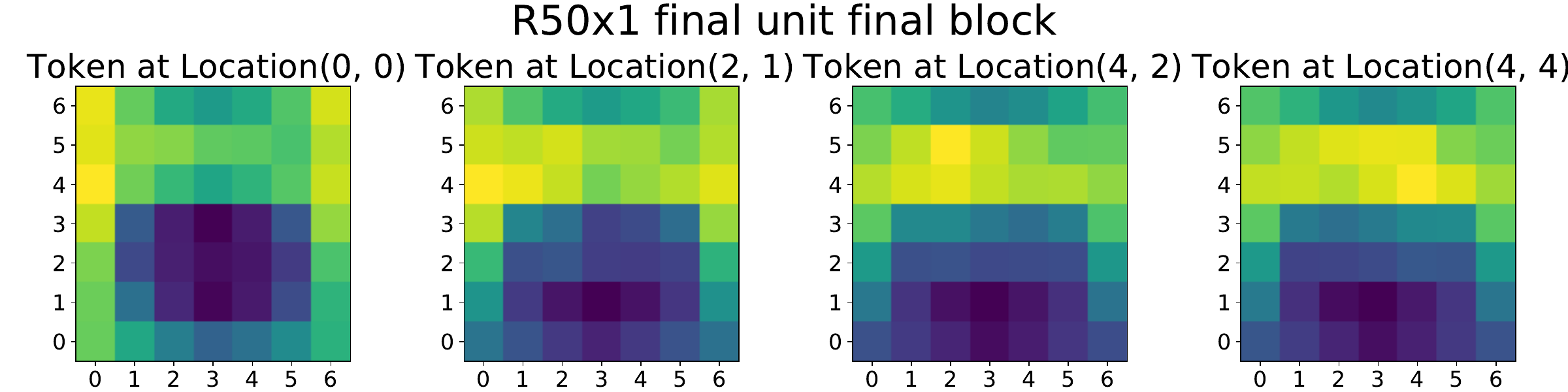} \\
    \caption{\small \textbf{Higher layers of ViT maintain spatial location information more faithfully than ResNets}. Each heatmap plot shows the CKA similarity between a single token representation in final block of the model and the input images, which are divided into non-overlapping patches. We observe that ViT tokens have strongest similarity to their corresponding spatial location in the image, but tokens corresponding to spatial locations at the edge of the image (e.g. token 0) additionally show similarity to other edge positions. This demonstrates that spatial information from the input is preserved even at the final layer of ViT. By contrast, ResNet ``tokens'' (features at a specific spatial location) are much less spatially discriminative, showing comparable similarity across a broad set of input spatial locations. See Appendix for additional layers and results.}
    \label{fig:localization-heatmaps}
\end{figure}

We begin by comparing token representations in the higher layers of ViT and ResNet to those of input patches. Recall that ViT tokens have a corresponding input patch, and thus a corresponding input spatial location. For ResNet, we define a token representation to be all the convolutional channels at a particular spatial location. This also gives it a corresponding input spatial location. We can then take a token representation and compute its CKA score with input image patches at different locations. The results are illustrated for different tokens (with their spatial locations labelled) in Figure \ref{fig:localization-heatmaps}.

For ViT, we observe that tokens corresponding to locations at the edge of the image are similar to edge image patches, but tokens corresponding to interior locations are well localized, with their representations being most similar to the corresponding image patch. By contrast, for ResNet, we see significantly weaker localization (though Figure \ref{fig:localization-heatmaps-resnet-app} shows improvements for earlier layers.)

\begin{figure}[t]
    \centering
    \includegraphics[width=0.8\columnwidth]{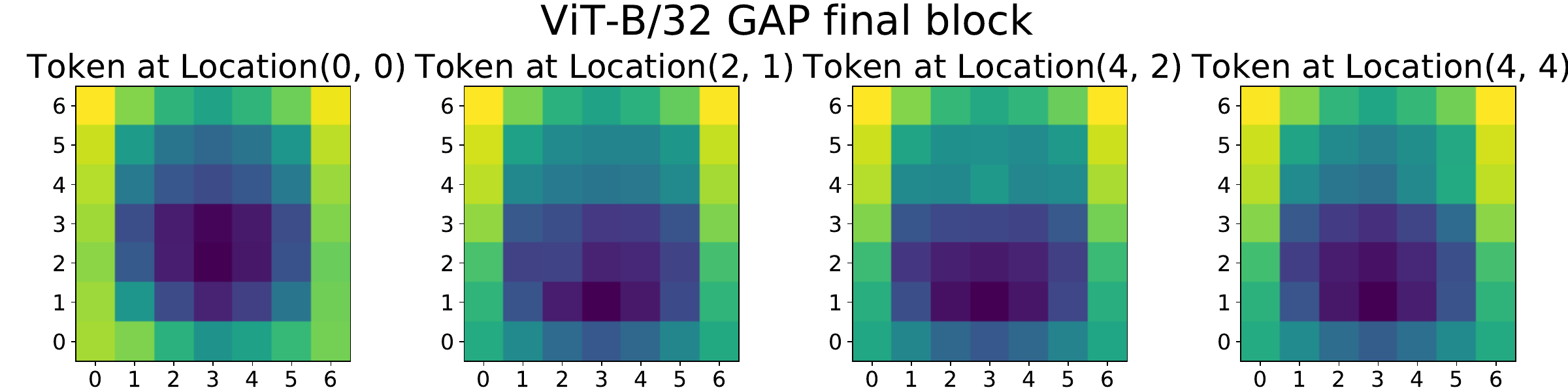} 
    \caption{\small \textbf{When trained with global average pooling (GAP) instead of a CLS token, ViTs show less clear localization (compare Figure \ref{fig:localization-heatmaps}).} We plot the same CKA heatmap between a token and different input images patches as in Figure \ref{fig:localization-heatmaps}, but for a ViT model trained with global average pooling (like ResNet) instead of a CLS token. We observe significantly less localization.}
    \label{fig:localization-heatmaps-gap}
    \vspace{-1em}
\end{figure}

One factor influencing this clear difference between architectures is that ResNet is trained to classify with a global average pooling step, while ViT has a separate classification (CLS) token. To examine this further, we test a ViT architecture trained with global average pooling (GAP) for localization (see Appendix~\ref{app:experimental_setup} for training details). The results, shown in Figure \ref{fig:localization-heatmaps-gap}, demonstrate that global average pooling does indeed reduce localization in the higher layers. More results in Appendix Section \ref{sec:localization-app}.

\begin{figure}[t]
    \centering
    \footnotesize
    \begin{tabular}{cc}
    \includegraphics[height=0.35\columnwidth]{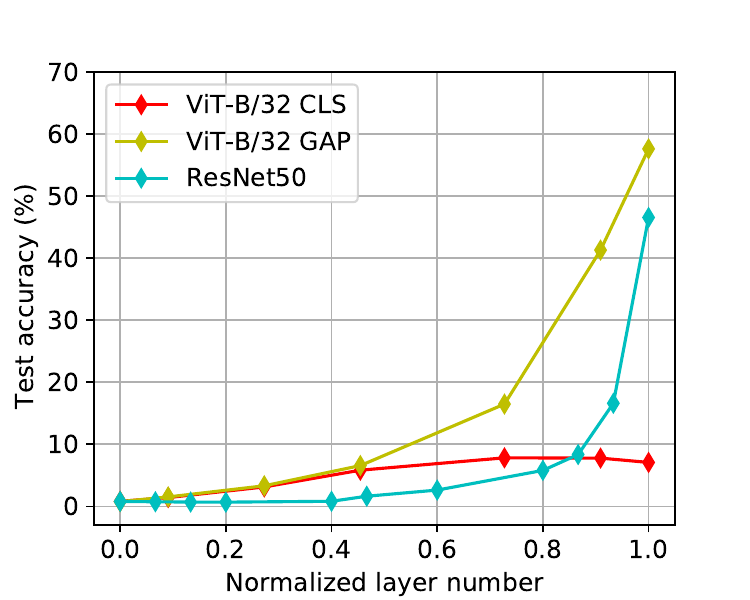} &
      \includegraphics[height=0.35\columnwidth]{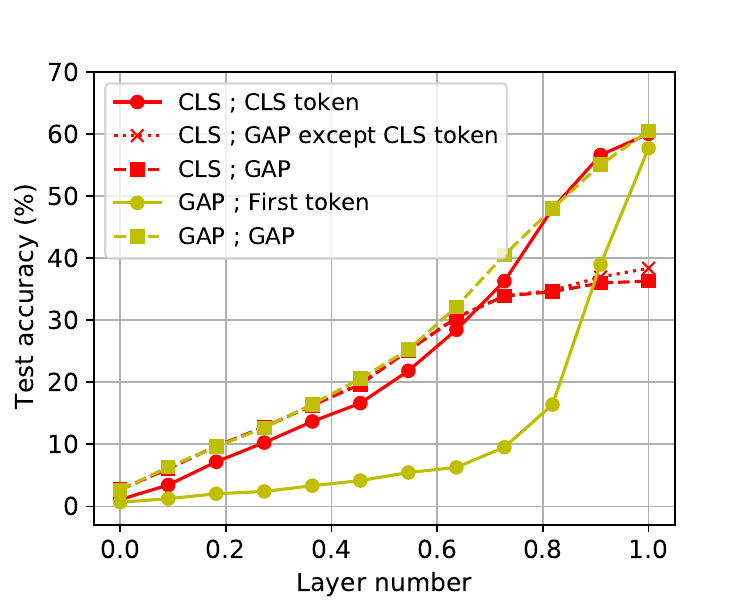} \\
      (a) Individual token evaluation &
      (b) CLS vs GAP models  \\
    \end{tabular}
    \caption{\textbf{Spatial localization experiments with linear probes.} We train linear classifiers on 10-shot ImageNet classification from the representations extracted from different layers of ViT-B/32 models. We then plot the accuracy of the probe versus the (normalized) layer number. \textbf{Left:} We train a classifier on each token separately and report the average accuracy over all tokens (excluding the CLS token for the ViT CLS model.) \textbf{Right:} Comparison of ViT models pre-trained with a classification token or with global average pooling (GAP) and then evaluated with different ways of aggregating the token representations.}
    \label{fig:linear_probes-2}
    \vspace{-1em}
\end{figure}

\textbf{Localization and Linear Probe Classification:}
The previous results have looked at localization through direct comparison of each token with input patches. To complete the picture, we look at using each token separately to perform \textit{classification} with linear probes. We do this across different layers of the model, training linear probes to classify image label with closed-form few-shot linear regression similar to~\citet{dosovitskiy2020image} (details in Appendix~\ref{app:experimental_setup}). Results are in Figure \ref{fig:linear_probes-2}, with further results in Appendix~\ref{sec:app-linear-probes}.
The left pane shows average accuracy of classifiers trained on individual tokens, where we see that ResNet50 and ViT with GAP model tokens perform well at higher layers, while in the standard ViT trained with a CLS token the spatial tokens do poorly -- likely because their representations remain spatially localized at higher layers, which makes global classification challenging. Supporting this are results on the right pane, which shows that a single token from the ViT-GAP model achieves comparable accuracy in the highest layer to all tokens pooled together. With the results of Figure \ref{fig:localization-heatmaps}, this suggests all higher layer tokens in GAP models learn similar (global) representations.

\vspace{-0.75em}
\section{Effects of Scale on Transfer Learning}
\vspace{-0.25em}
\label{sec:transfer-learning}
\begin{figure}
    \centering
    \includegraphics[width=0.85\columnwidth]{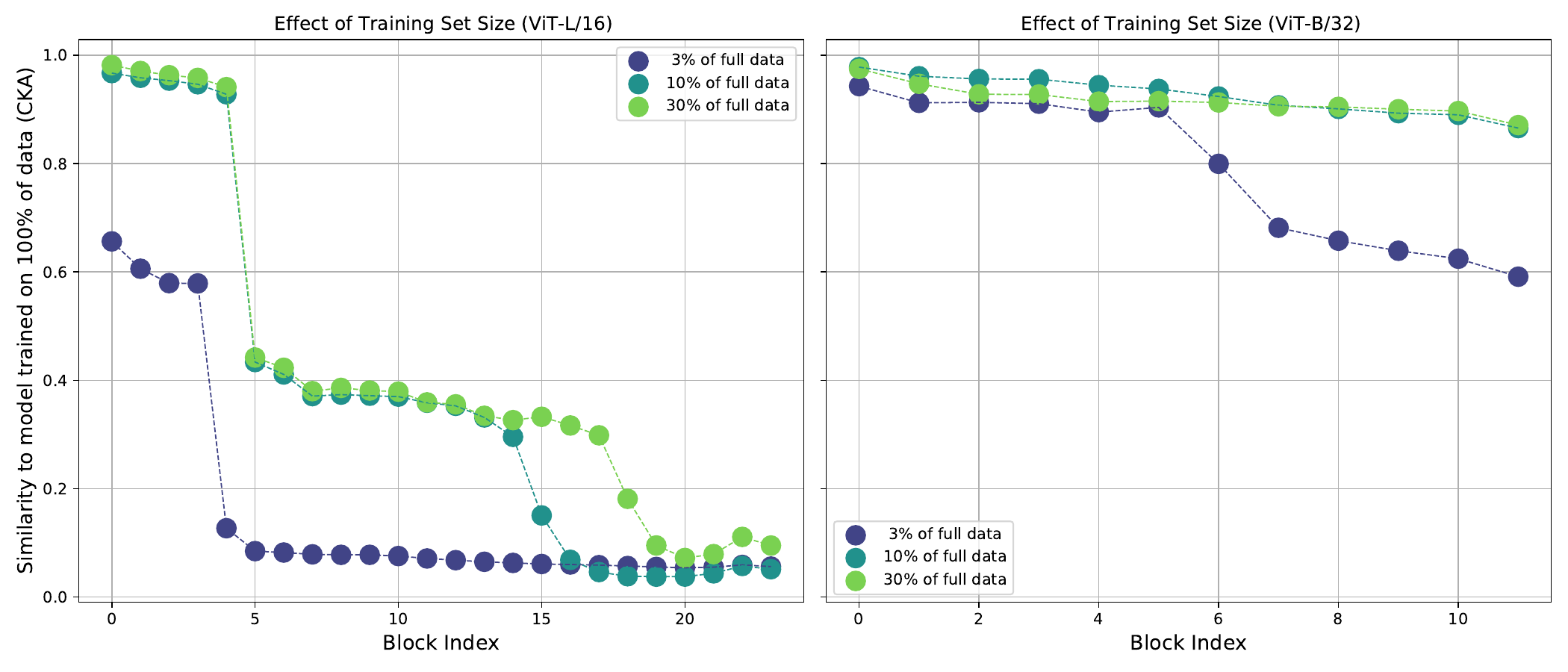} 
    \vspace{-0.5em}
    \caption{\small \textbf{Measuring similarity of representations learned with varying amounts of data shows the importance of large datasets for higher layers and larger model representations.} We compute the similarity of representations at each block for ViT models that have been trained on smaller subsets of the data to a model that has been trained on the full data on ViT-L/16 (left) and ViT-B/32 (right). We observe that while lower layer representations have high similarity even with ~$10\%$ of the data, higher layers and larger models require significantly more data to learn similar representations.}
    \label{fig:training-set-size}
    \vspace{-1em}
\end{figure}
Motivated by the results of \citet{dosovitskiy2020image} that demonstrate the importance of dataset scale for high performing ViTs, and our earlier result (Figure~\ref{fig:attention-heads-scale}) on needing scale for local attention, we perform a study of the effect of dataset scale on representations in transfer learning.

We begin by studying the effect on representations as the JFT-300M pretraining dataset size is varied. Figure \ref{fig:training-set-size} illustrates the results on ViT-B/32 and ViT-L/16. Even with $3\%$ of the entire dataset, lower layer representations are very similar to the model trained on the whole dataset, but higher layers require larger amounts of pretraining data to learn the same representations as at large data scale, especially with the large model size. In Section \ref{sec:app-scale-transfer-learning}, we study how much representations change in finetuning, finding heterogeneity over datasets.  

\begin{figure}
    \centering
    \footnotesize
    \begin{tabular}{cc}
    \includegraphics[height=0.35\columnwidth]{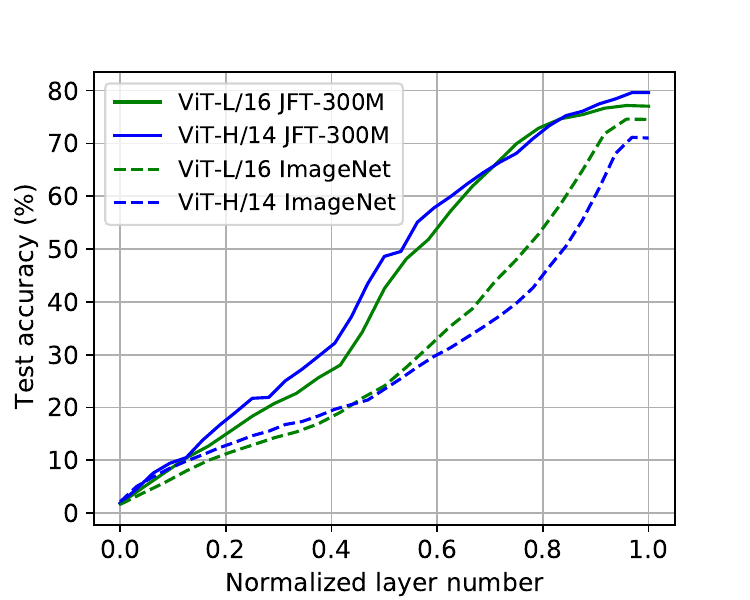} &
    \includegraphics[height=0.35\columnwidth]{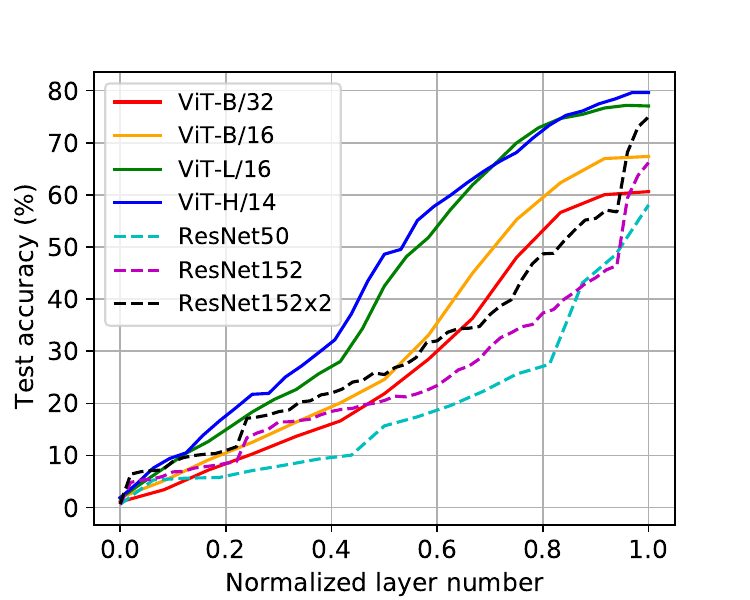} \\
    (a) JFT-300M vs ImageNet pre-training &
    (b) ViTs vs ResNets
    \end{tabular}
    \caption{\small Experiments with linear probes. We train linear classifiers on 10-shot ImageNet classification from the aggregated representations of different layers of different models. We then plot the accuracy of the probe versus the (normalized) layer number. \textbf{Left:} Comparison of ViTs pre-trained on JFT-300M or ImageNet and evaluated with linear probes on Imagenet.
    \textbf{Right:} Comparison of ViT and ResNet models trained JFT-300m, evaluated with linear probes on ImageNet.}
    \label{fig:linear_probes-1}
    \vspace{-1em}
\end{figure}

We next look at dataset size effect on the larger ViT-L/16 and ViT-H/14 models. Specifically, in the left pane of Figure \ref{fig:linear_probes-1}, we train linear classifer probes on ImageNet classes for models pretrained on JFT-300M vs models only pretrained on ImageNet. We observe the JFT-300M pretained models achieve much higher accuracies even with middle layer representations, with a ~$30\%$ gap in absolute accuracy to the models pretrained only on ImageNet. This suggests that for larger models, the larger dataset is especially helpful in learning high quality intermediate representations. This conclusion is further supported by the results of the right pane of Figure \ref{fig:linear_probes-1}, which shows linear probes on different ResNet and ViT models, all pretrained on JFT-300M. We again see the larger ViT models learn much stronger intermediate representations than the ResNets. Additional linear probes experiments in Section \ref{sec:app-linear-probes} demonstrate this same conclusion for transfer to CIFAR-10 and CIFAR-100. 

\section{Discussion}
\label{sec:discussion}
\textbf{Limitations:} Our study uses CKA \cite{kornblith2019similarity}, which summarizes measurements into a single scalar, to provide quantitative insights on representation similarity. While we have complemented this with interventional tests and other analyses (e.g. linear probes), more fine-grained methods may reveal additional insights and variations in the representations. 

\textbf{Conclusion:} 
Given the central role of convolutional neural networks in computer vision breakthroughs, it is remarkable that Transformer architectures (almost \textit{identical} to those used in language) are capable of similar performance. This raises fundamental questions on whether these architectures work in the same way as CNNs. Drawing on representational similarity techniques, we find surprisingly clear differences in the features and internal structures of ViTs and CNNs. An analysis of self-attention and the strength of skip connections demonstrates the role of earlier global features and strong representation propagation in ViTs for these differences, while also revealing that some CNN properties, e.g. local information aggregation at lower layers, are important to ViTs, being learned from scratch at scale. We examine the potential for ViTs to be used beyond classification through a study of spatial localization, discovering ViTs with CLS tokens show strong preservation of spatial information --- promising for future uses in object detection. Finally, we investigate the effect of scale for transfer learning, finding larger ViT models develop significantly stronger intermediate representations through larger pretraining datasets. These results are also very pertinent to understanding MLP-based architectures for vision proposed by concurrent work \cite{tolstikhin2021mlp, touvron2021resmlp}, further discussed in Section \ref{sec:mlp-mixer}, and together answer central questions on differences between ViTs and CNNs, and suggest new directions for future study. From the perspective of societal impact, these findings and future work may help identify potential failures as well as greater model interpretability.   

\bibliography{references.bib}

\clearpage
\counterwithin{figure}{section}
\counterwithin{table}{section}
\appendix
\part*{Appendix}
Additional details and results from the different sections are included below.

\section{Additional details on Methods and the Experimental Setup}
\label{app:experimental_setup}
To understand systematic differences between ViT and CNNs, we use a representative set of different models of each type, guided by the performance results in \cite{dosovitskiy2020image}. Specifically, for ViTs, we look at ViT-B/32, ViT-L/16 and ViT-H/14, where the smallest model (ViT-B/32) shows limited improvements when pretraining on JFT-300M \cite{sun2017revisiting} vs. the ImageNet Large Scale Visual Recognition Challenge 2012 dataset \cite{deng2009imagenet,russakovsky2015imagenet}, while the largest, ViT-H/14, achieves state of the art when pretrained on JFT-300M \cite{sun2017revisiting}. ViT-L/16 is close to the performance ViT-H/14 \cite{dosovitskiy2020image}. For CNNs, we look at ResNet50x1 which also shows saturating performance when pretraining on JFT-300M, and also ResNet152x2, which in contrast shows large performance gains with increased pretraining dataset size. As in \citet{dosovitskiy2020image}, these ResNets follow some of the implementation changes first proposed in BiT \cite{kolesnikov2019big}. 

In addition to the standard Vision Transformers trained with a classification token (CLS), we also trained ViTs with global average pooling (GAP).
In these, there is no classification token~-- instead, the representations of tokens in the last layer of the transformer are averaged and directly used to predict the logits.
The GAP model is trained with the same hyperparameters as the CLS one, except for the initial learning rate that is set to a lower value of $0.0003$.

For analyses of internal model representations, we observed no meaningful difference between representations of images drawn from ImageNet and images drawn from JFT-300M. Figures~\ref{fig:representation-structure}, \ref{fig:representation-cross-comparisons}, \ref{fig:localization-heatmaps}, and \ref{fig:localization-heatmaps-gap} use images from the JFT-300M dataset that were not seen during training, while Figures~\ref{fig:receptive-fields-summary}, \ref{fig:attention-heads-plot}, \ref{fig:attention-heads-scale}, \ref{fig:local-global-split} \ref{fig:identity-connection-norms-vit}, \ref{fig:identity-connection-norms-cka}, and \ref{fig:training-set-size} use images from the ImageNet 2012 validation set. Figures~\ref{fig:linear_probes-2} and \ref{fig:linear_probes-1} involve 10-shot probes trained on the ImageNet 2012 training set, tuned hyperparameters on a heldout portion of the training set, and evaluated on the validation set.

\paragraph{Additional details on CKA implementation} To compute CKA similarity scores, we use minibatch CKA, introduced in \cite{nguyen2020wide}. Specifically, we use batch sizes of $1024$ and we sample a total of $10240$ examples without replacement. We repeat this $20$ times and take the average. Experiments varying the exact batch size (down to $128$), and total number of examples used for CKA ( down to $2560$ total examples, repeated $10$ times), had no noticeable effect on the results. 

\paragraph{Additional details on linear probes.}
We train linear probes as regularized least-squares regression, following~\citet{dosovitskiy2020image}.
We map the representations training images to $\{-1, 1\}^N$ target vectors, where $N$ is the number of classes.
The solution can be recovered efficiently in closed form.

For vision transformers, we train linear probes on representations from individual tokens or on the representation averaged over all tokens, at the output of different transformer layers (each layer meaning a full transformer block including self-attention and MLP). 
For ResNets, we take representation at the output of each residual block (including 3 convolutional layers). The resolution of the feature maps changes throughout the model, so we perform an additional pooling step bringing the feature map to the same spatial size as in the last stage.
Moreover, ResNets differ from ViTs in that the number of channels changes throughout the model, with fewer channels in the earlier layers. 
This smaller channel count in the earlier layers could potentially lead to worse performance of the linear probes.
To compensate for this, before pooling we split the feature map into patches and flattened each patch, so as to arrive at the channel count close to the channel count in the final block.
All results presented in the paper include this additional patching step; however, we have found that it brings only a very minor improvement on top of simple pooling.

\FloatBarrier

\section{Additional Representation Structure Results}
Here we include some more CKA heatmaps, which provide insights on model representation structures (compare to Figure \ref{fig:representation-structure}, Figure \ref{fig:representation-cross-comparisons} in the main text.) We observe similar conclusions: ViT representation structure has a more uniform similarity structure across layers, and comparing ResNet to ViT representations show a large fraction of early ResNet layers similar to a smaller number of ViT layers.
\begin{figure}[h]
    \centering
    \includegraphics[width=0.5\columnwidth]{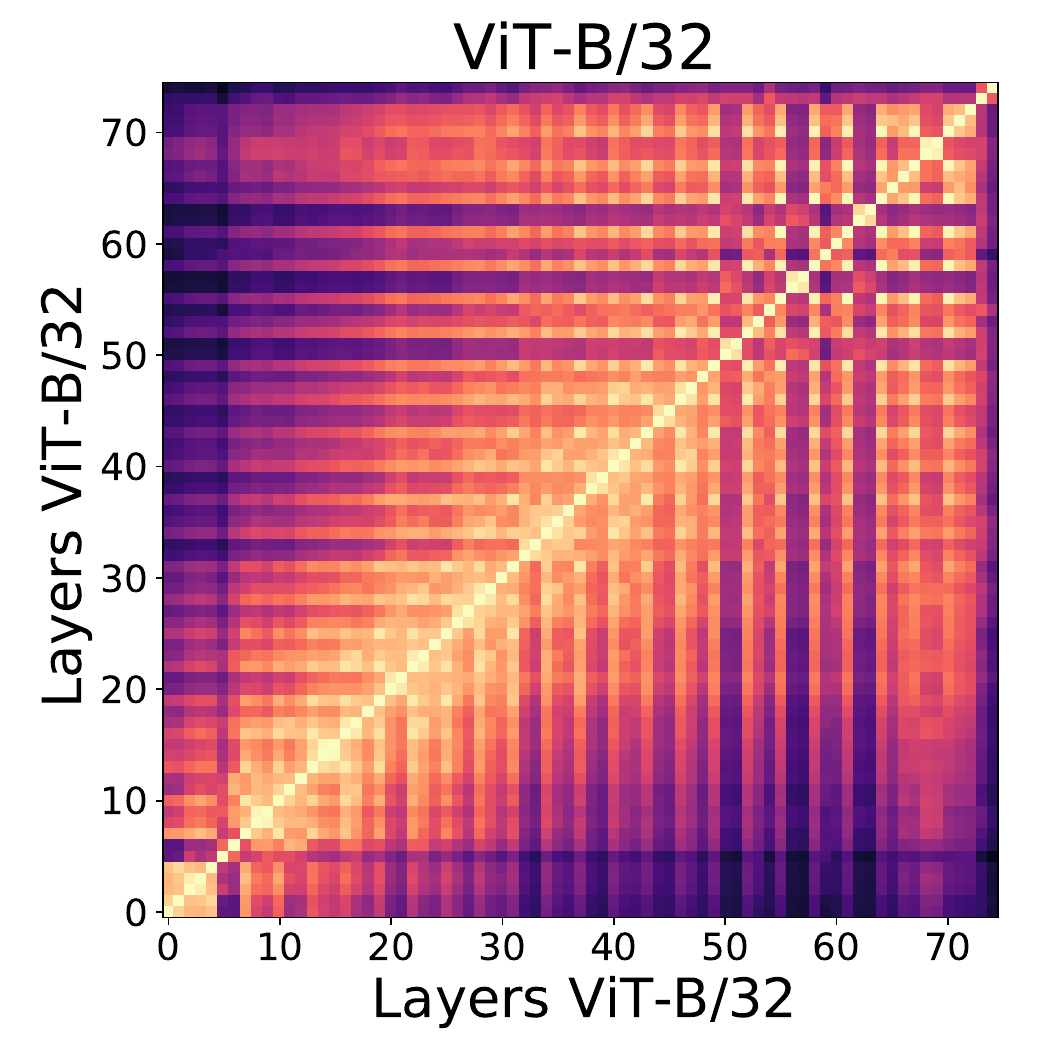}
    \begin{tabular}{cc}
     \includegraphics[width=0.47\columnwidth]{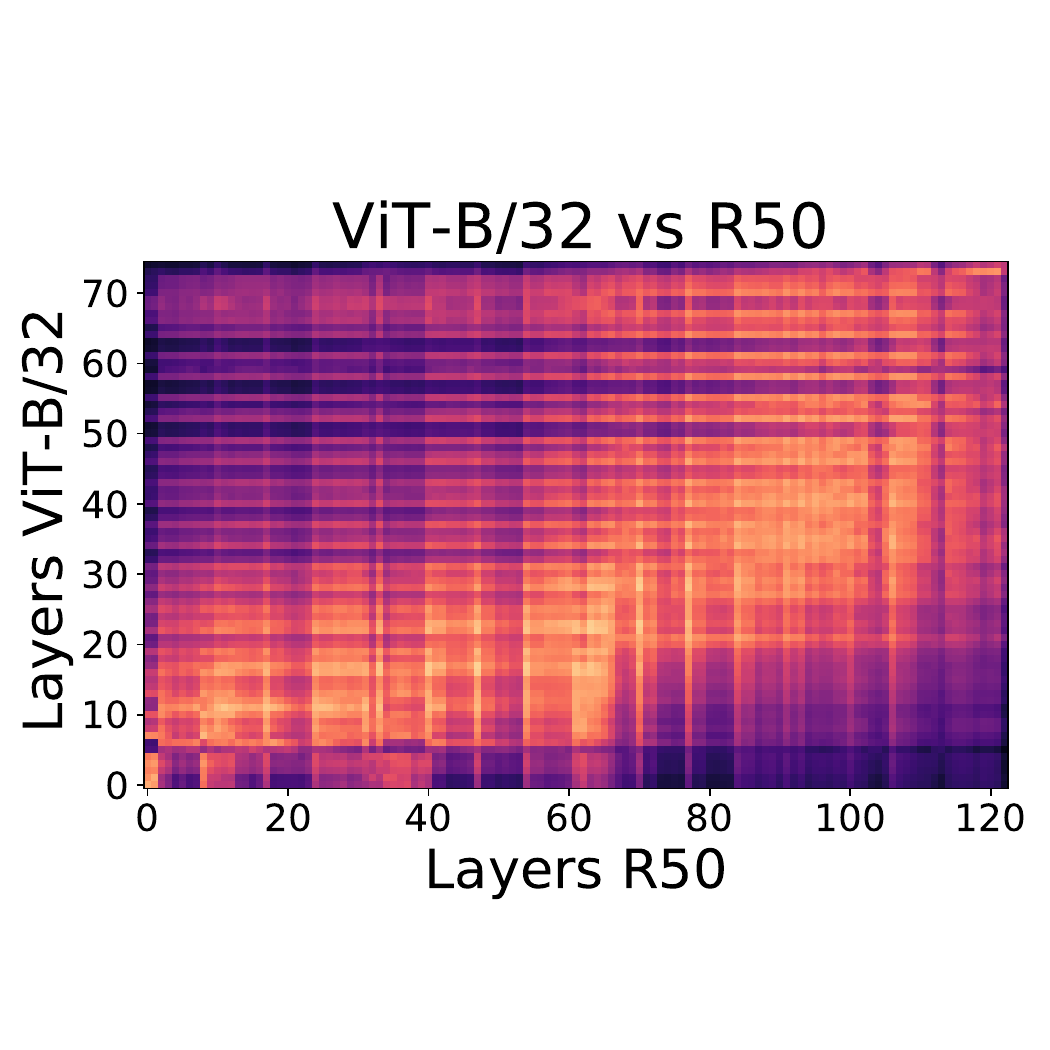} &
     \includegraphics[width=0.5\columnwidth]{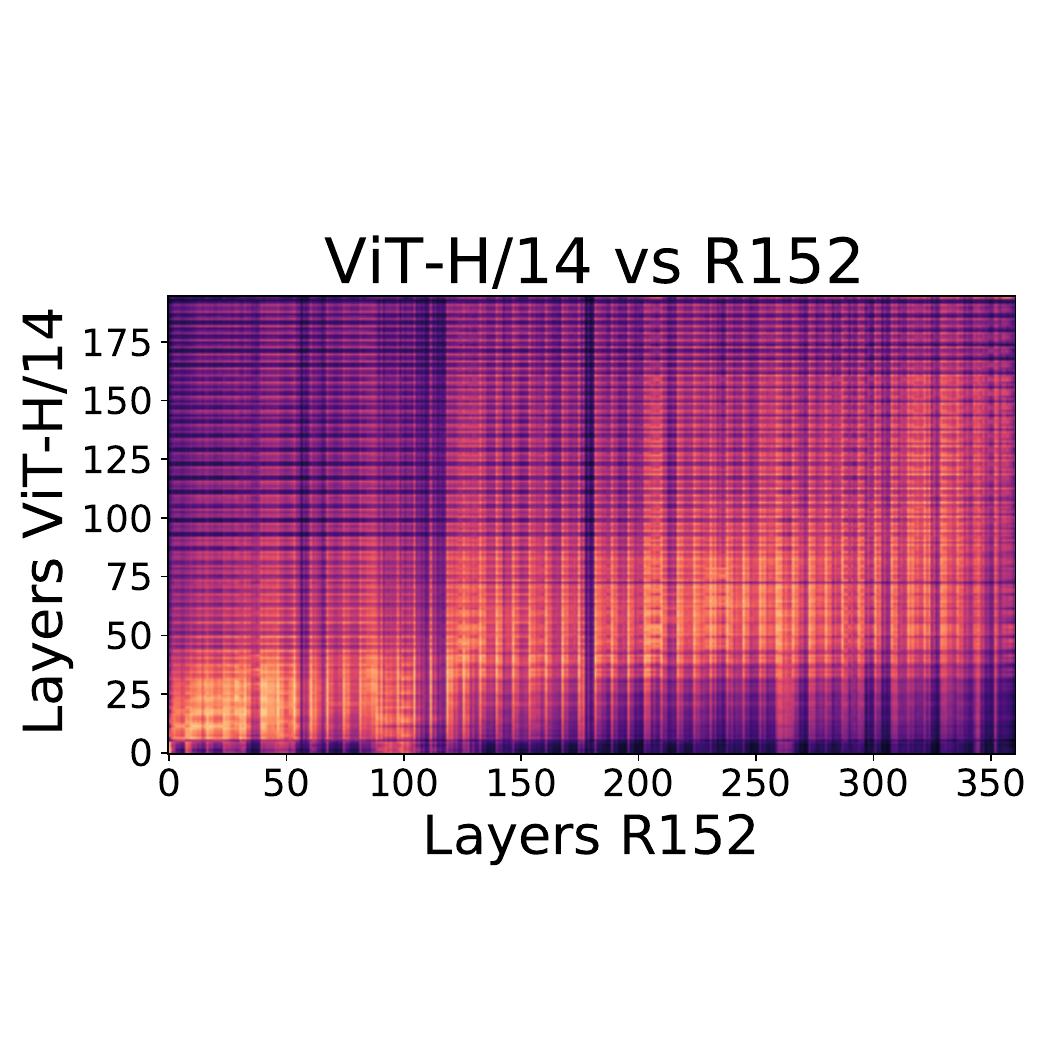} \\
    \end{tabular}
    \caption{\small \textbf{Additional CKA heatmap results.} Top shows CKA heatmap for ViT-B/32, where we can also observe strong similarity between lower and higher layers and the grid like, uniform representation structure. Bottom shows (i) ViT-B/32 compared to R50, where we again see that ~60 of the lowest R50 layers are similar to about ~25 of the lowest ViT-B/32 layers, with the remaining layers most similar to each other (ii) ViT-H/14 compared to R152, where we see the lowest ~100 layers of R152 are most similar to lowest ~30 layers of ViT-H/14.}
    \label{fig:representation-heatmap-app}
\end{figure}
\FloatBarrier

\section{Additional Local/Global Information Results}
\begin{figure}
    \centering
    \includegraphics[width=\linewidth]{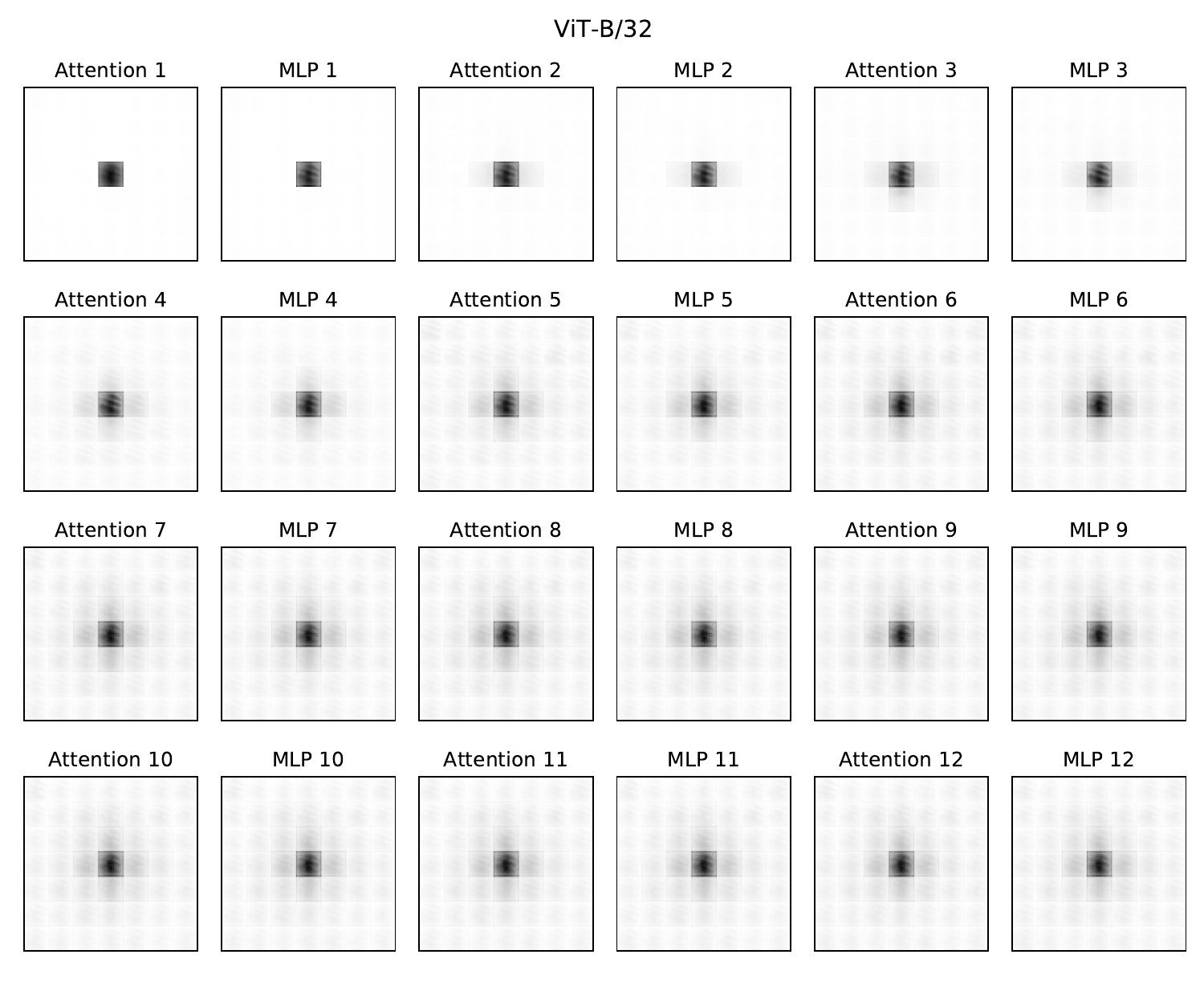}\\
    \vskip -0.5em
    \caption{\small \textbf{Post-residual receptive fields of all ViT-B/32 sublayers.}}
    \label{fig:receptive-fields-vitb-full}
\end{figure}

\begin{figure}
    \centering
    \includegraphics[width=\linewidth]{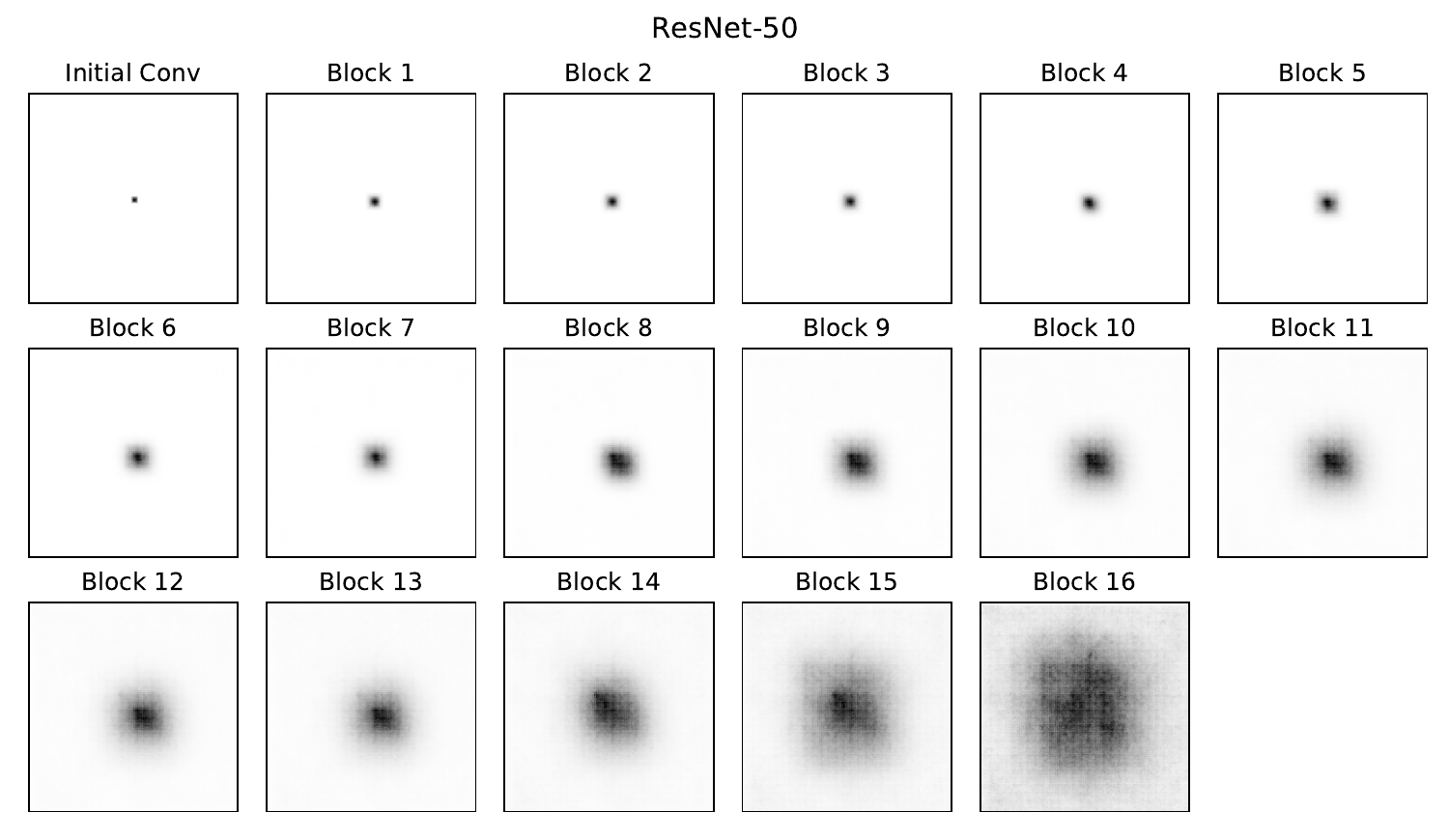}\\
    \vskip -0.5em
    \caption{\small \textbf{Post-residual receptive fields of all ResNet-50 blocks.}}
    \label{fig:receptive-fields-r50-full}
\end{figure}

\begin{figure}
    \centering
    \includegraphics[width=\linewidth]{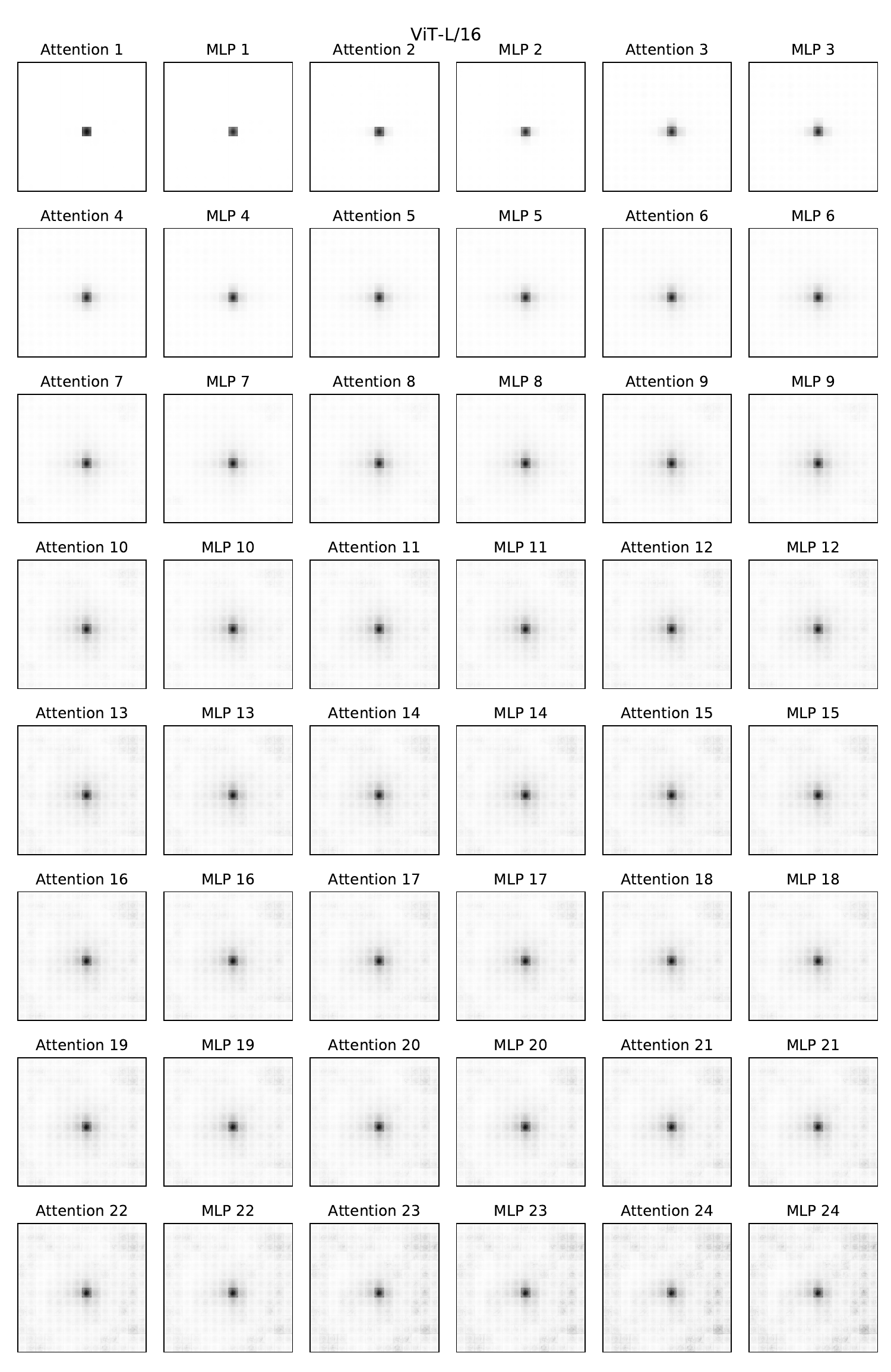}\\
    \vskip -0.5em
    \caption{\small \textbf{Post-residual receptive fields of all ViT-L/16 sublayers.}}
    \label{fig:receptive-fields-vitl-full}
\end{figure}

\begin{figure}
    \centering
    \includegraphics[width=\linewidth]{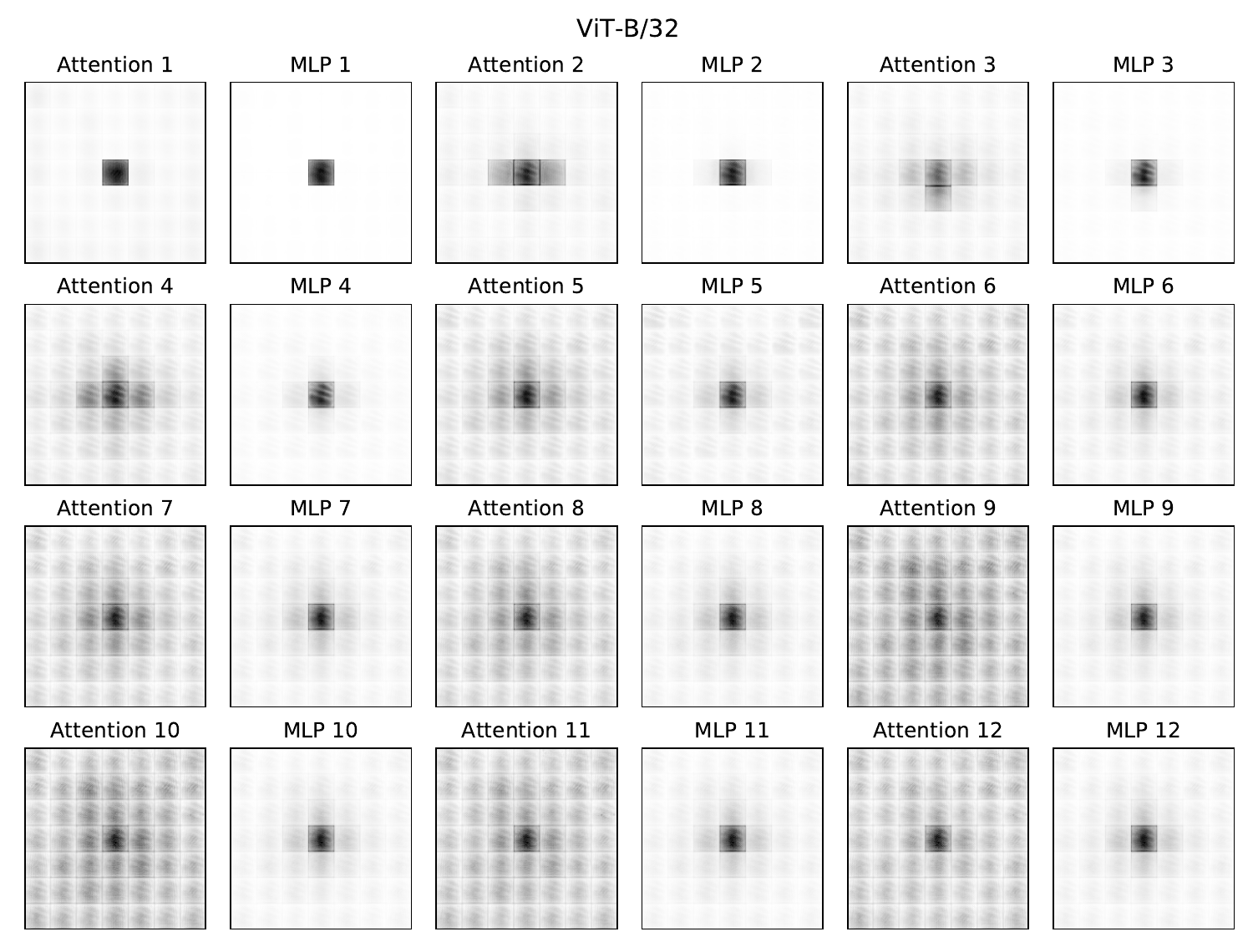}\\
    \includegraphics[width=\linewidth]{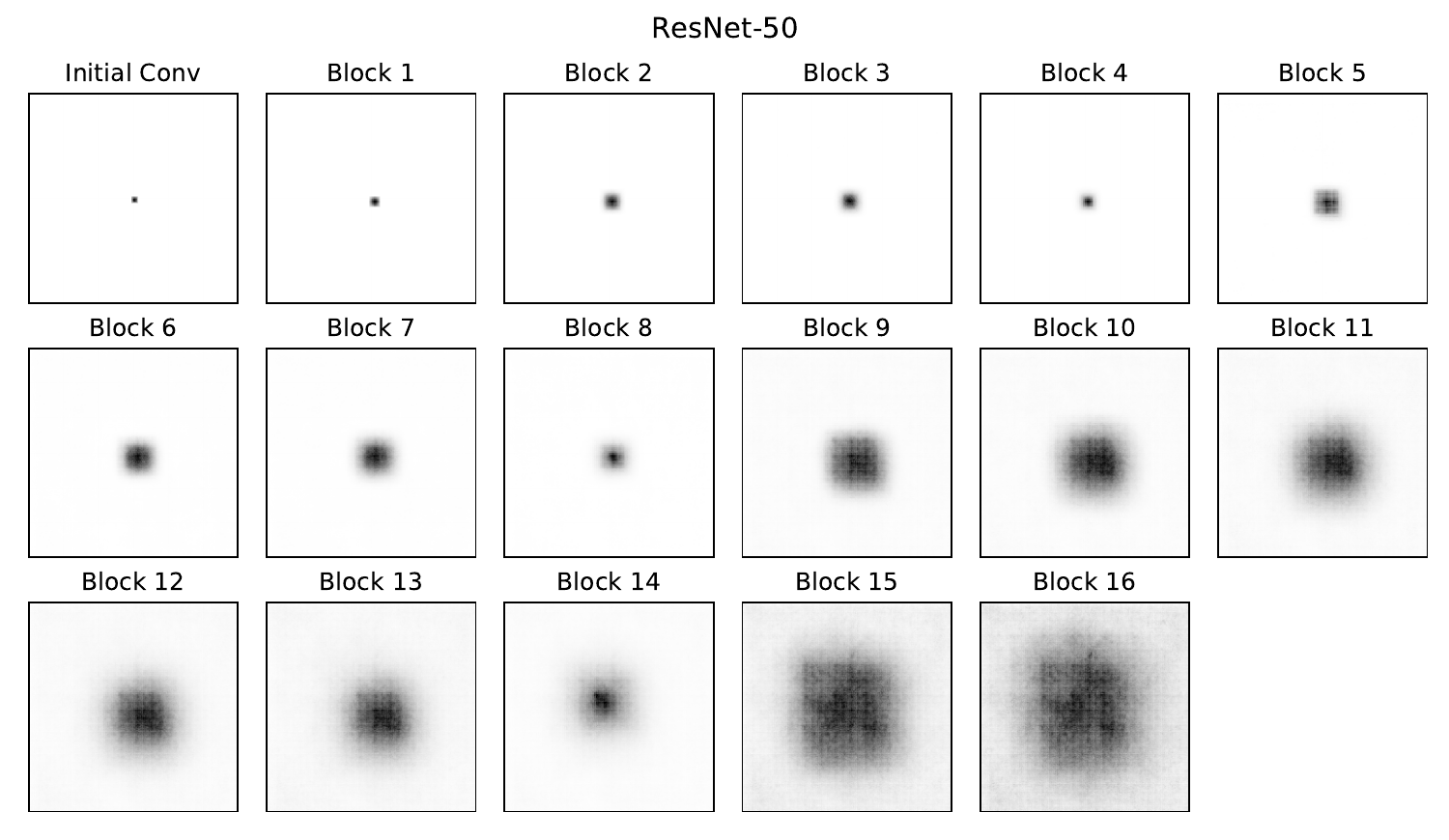}\\
    \vskip -0.5em
    \caption{\small \textbf{Pre-residual receptive fields of all ViT-B/32 sublayers and ResNet-50 blocks.} In ViT-B, we see that the pre-residual receptive fields of later attention sublayers are not dominated by the center patch, in contrast to the post-residual receptive fields shown in  Figure~\ref{fig:receptive-fields-vitb-full}. Thus, although later attention sublayers integrate information across the entire input image, network representations remain localized due to the strong skip connections. ResNet-50 pre-residual receptive fields generally resemble the post-residual receptive fields shown in Figure~\ref{fig:receptive-fields-r50-full}. The receptive field appears to ``shrink'' at blocks 4, 8, and 14, which are each the first in a stage, and for which we plot only the longer branch and not the shortcut. The receptive field does not shrink when computed after the summation of these branches, as shown in Figure~\ref{fig:receptive-fields-r50-full}.}
    \label{fig:receptive-fields-preresidual}
\end{figure}

In Figures~\ref{fig:receptive-fields-vitb-full}, ~\ref{fig:receptive-fields-r50-full}, and ~\ref{fig:receptive-fields-vitl-full}, we provide full plots of effective receptive fields of all layers of ViT-B/32, ResNet-50, and ViT-L/16, taken after the residual connections as in Figure~\ref{fig:receptive-fields-summary} in the text. In Figure~\ref{fig:receptive-fields-preresidual}, we show receptive fields of ViT-B/32 and ResNet-50 taken \textit{before} the residual connections. Although the pre-residual receptive fields of ViT MLP sublayers resemble the post-residual receptive fields in Figure~\ref{fig:receptive-fields-vitb-full}, the pre-residual receptive fields of attention sublayers have a smaller relative contribution from the corresponding input patch. These results support our findings in Section~\ref{sec:local-global} regarding the global nature of attention heads, but suggest that network representations remain tied to input patch locations because of the strong contributions from skip connections, studied in Section~\ref{sec:skip-connections}. ResNet-50 pre-residual receptive fields look similar to the post-residual receptive fields.

\label{app:local-global}
\FloatBarrier

\begin{figure}[h]
\centering
\begin{tabular}{cc}
     \includegraphics[width=0.5\columnwidth]{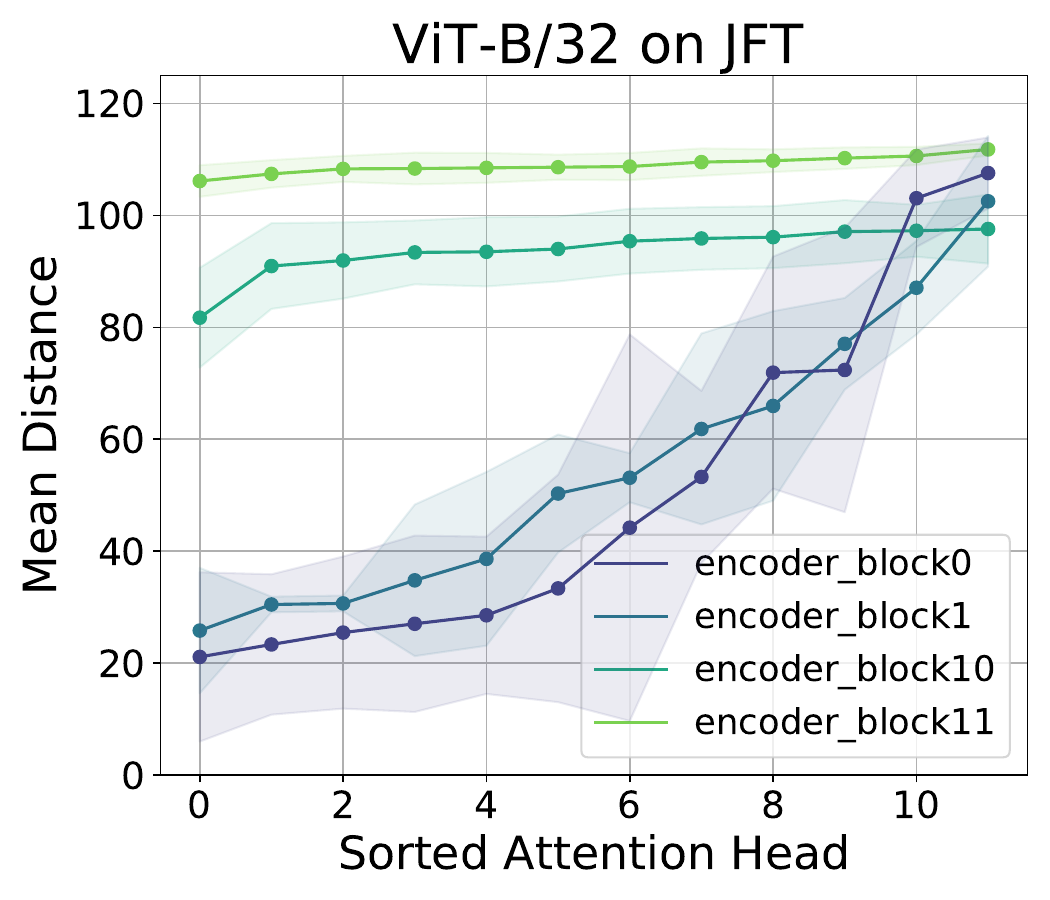} &
     \includegraphics[width=0.5\columnwidth]{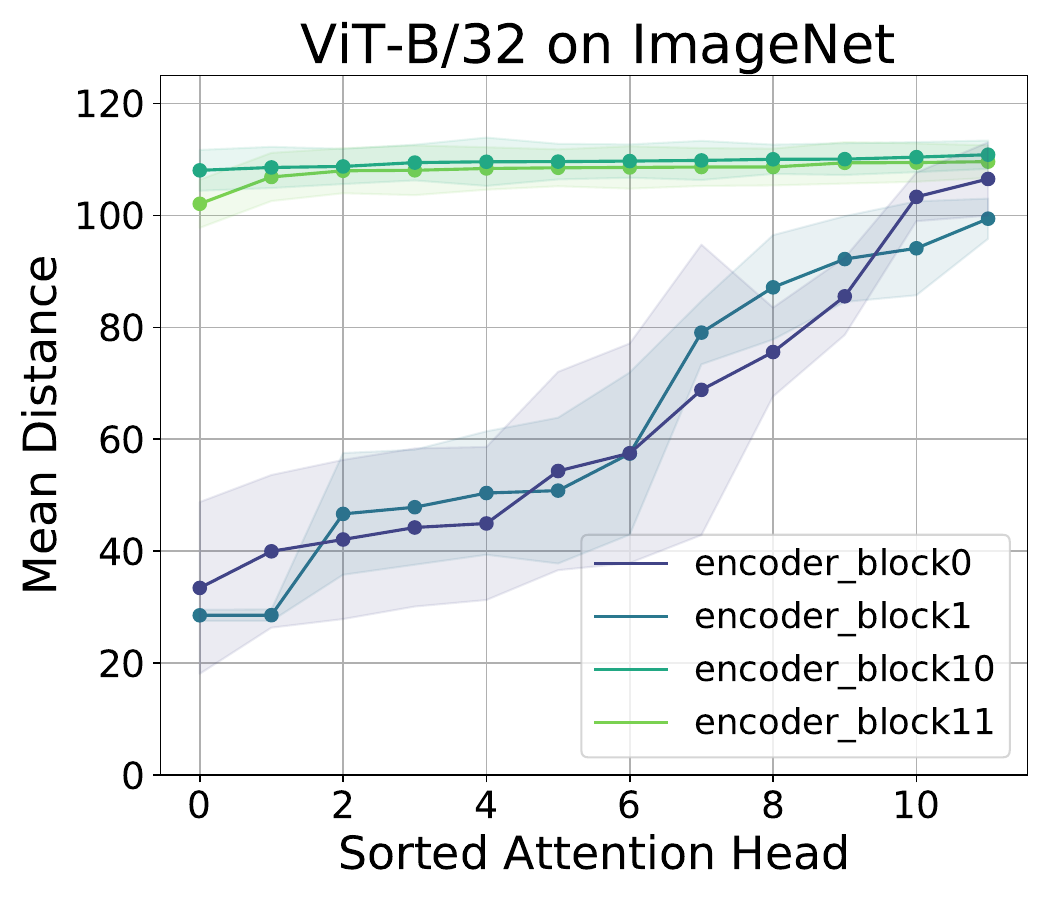} 
\end{tabular}
    \caption{\small \textbf{Plot of attention head distances for ViT-B/32 when trained on JFT-300M and on only ImageNet shows that ViT-B/32 learns to attend locally even on a smaller dataset.} Compare to Figure \ref{fig:attention-heads-plot} in the main text. While ViT-L and ViT-H show large performance improvements when  (i) finetuned on ImageNet having been pretrained on JFT compared to (ii) being trained on only ImageNet, ViT-B/32 has similar performance in both settings. We also observe that ViT-H, ViT-L don't learn to attend locally in the lowest layers when only trained on ImageNet (Figure \ref{fig:attention-heads-plot}), whereas here we see ViT-B/32 still learns to attend locally --- suggesting connections between performance and heads learning to attend locally.}
    \label{fig:attention-heads-ViTB-app}
\end{figure}

\begin{figure}[h]
\centering
    \includegraphics[width=0.8\columnwidth]{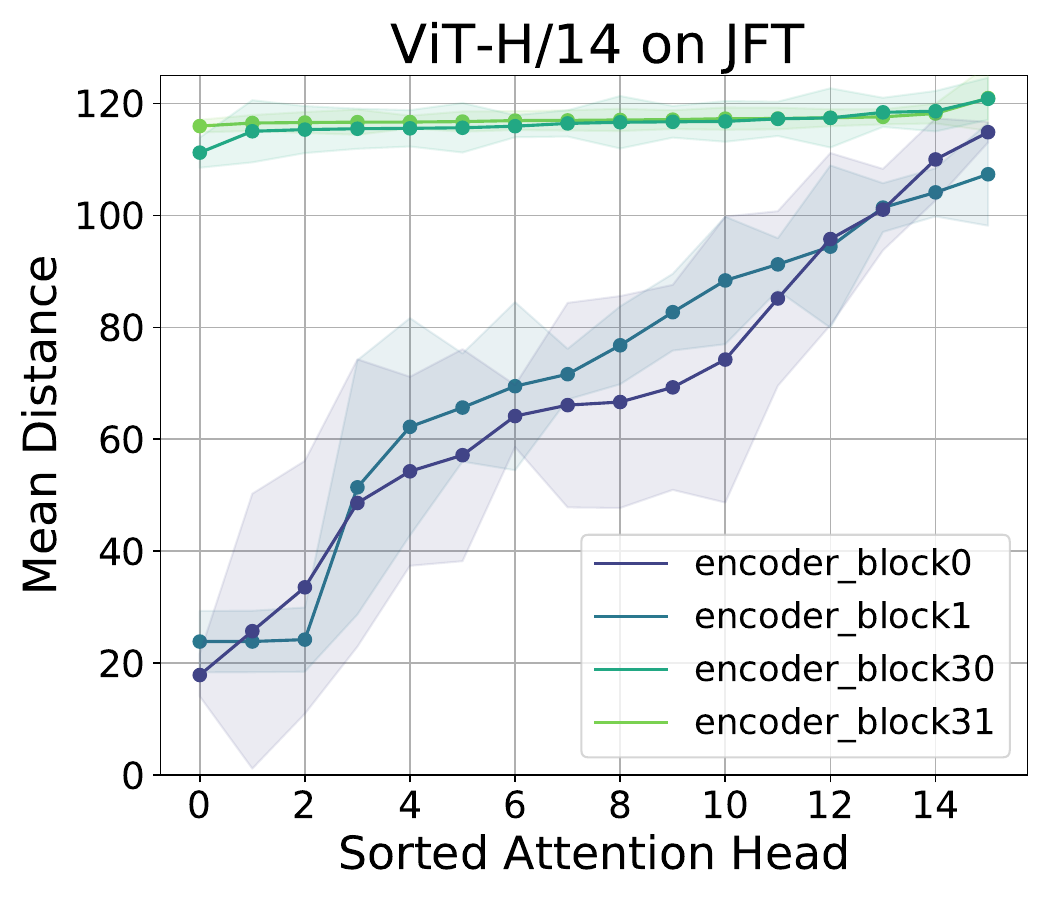} \\
    \caption{\small \textbf{Additional plot of attention head distances for ViT-H/14 on JFT-300M.} Compare to Figure \ref{fig:attention-heads-plot} in the main text. For each attention head, we compute the pixel distance it attends to, weighted by the attention weights, and then average over 5000 datapoints to get an average attention head distance. We plot the heads sorted by their average attention distance for the two lowest and two highest layers in the ViT, observing that the lower layers attend both locally and globally, while the higher layers attend entirely globally.}
    \label{fig:attention-heads-plot-app}
\end{figure}

\section{Localization}
\label{sec:localization-app}
Below we include additional localization results: computing CKA between different input patches and tokens in the higher layers of the models. We show results for ViT-H/14, additional higher layers for ViT-L/16, ViT-B/32 and addtional layers for ResNet.
\begin{figure}[h]
    \centering
    \includegraphics[width=\columnwidth]{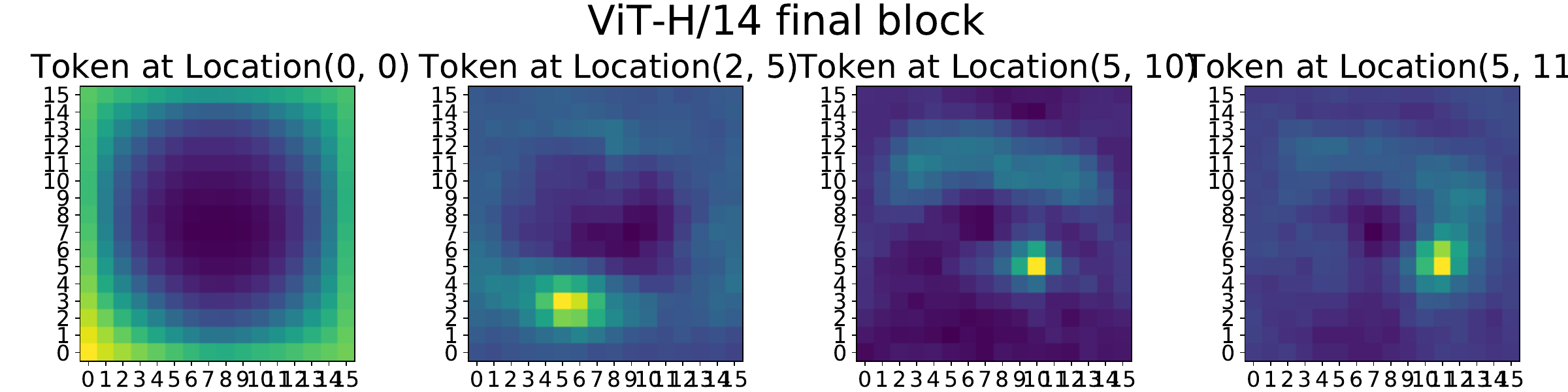} \\
     \includegraphics[width=\columnwidth]{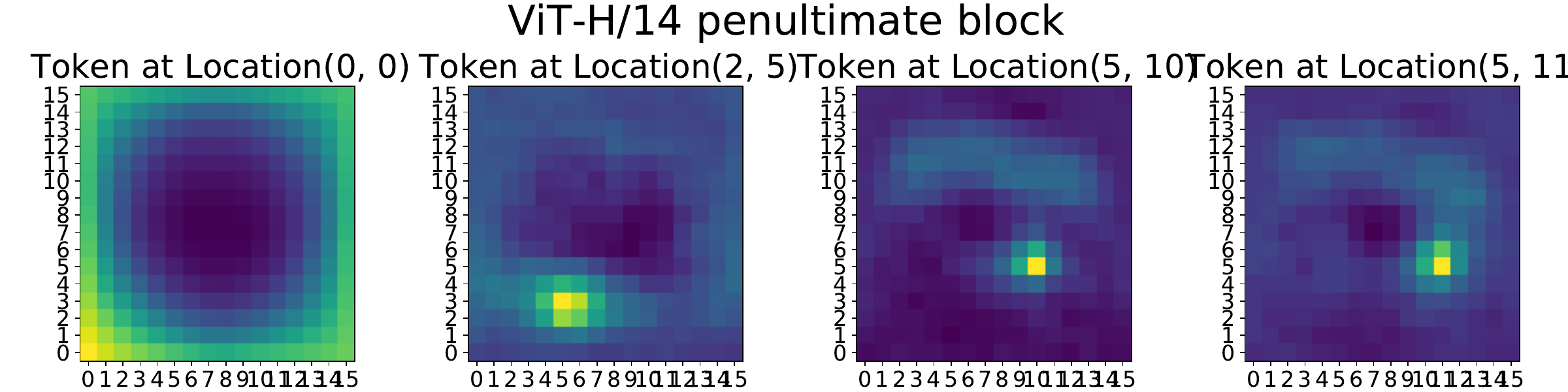} \\
    \caption{\small \textbf{Localization heatmaps for ViT-H/14}. We see that ViT-H/14 is also well localized, both in the final block and penultimate block, with tokens with corresponding locations in the interior of the image most similar to the image patches at those locations, while tokens on the edge are similar to many edge positions.}
    \label{fig:localization-heatmaps-vit-h14-app}
\end{figure}

\begin{figure}[h]
    \centering
    \includegraphics[width=\columnwidth]{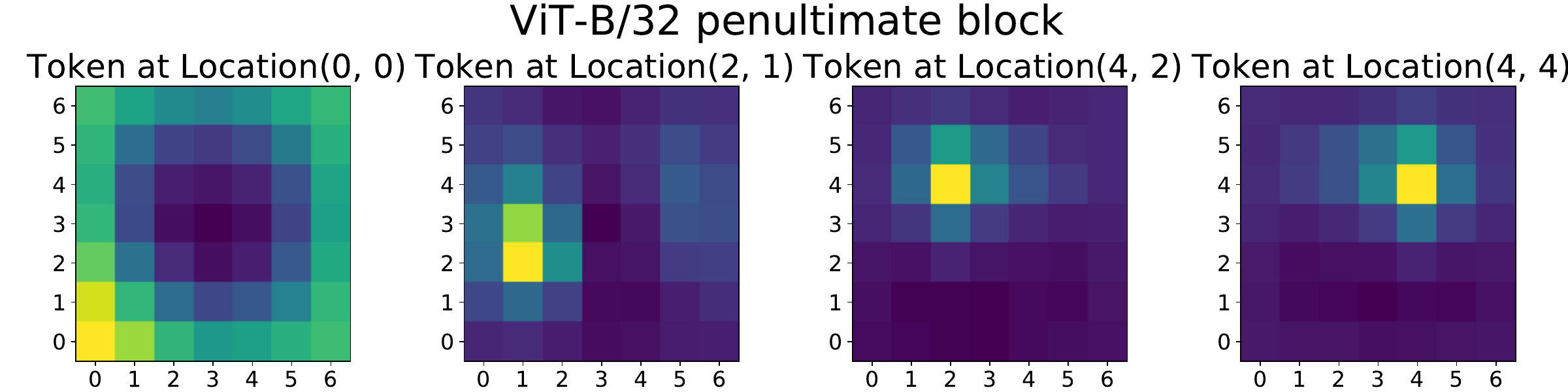} \\
    \includegraphics[width=\columnwidth]{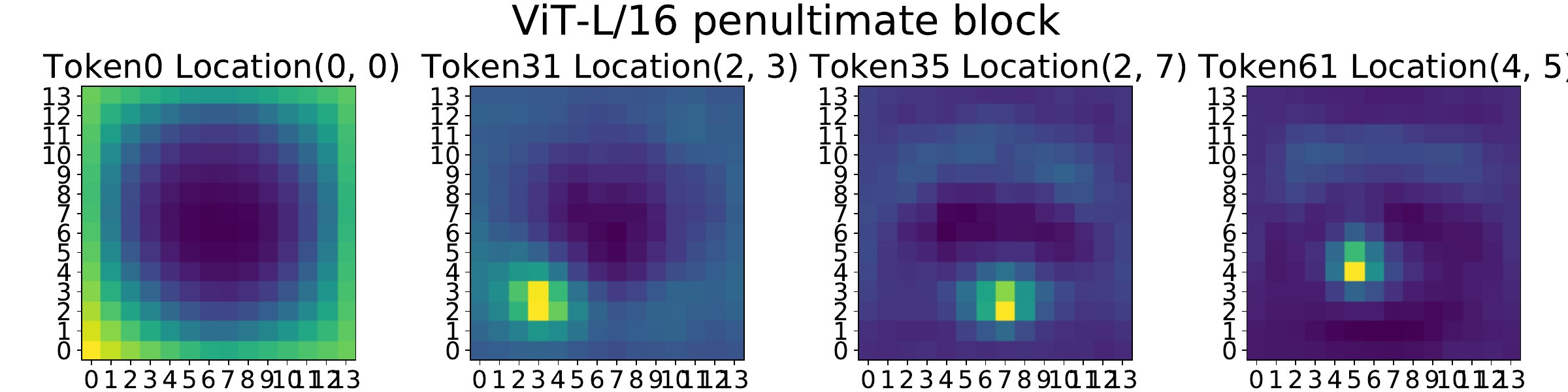} \\
    \caption{\small \textbf{Additional localization heatmaps for other higher layers of ViT-L/16 and ViT-B/32}. We see that models (as expected) remain well localized in higher layers other than the final block.}
    \label{fig:localization-heatmaps-vit-other-layers-app}
\end{figure}

\begin{figure}[h]
    \centering
   \includegraphics[width=\columnwidth]{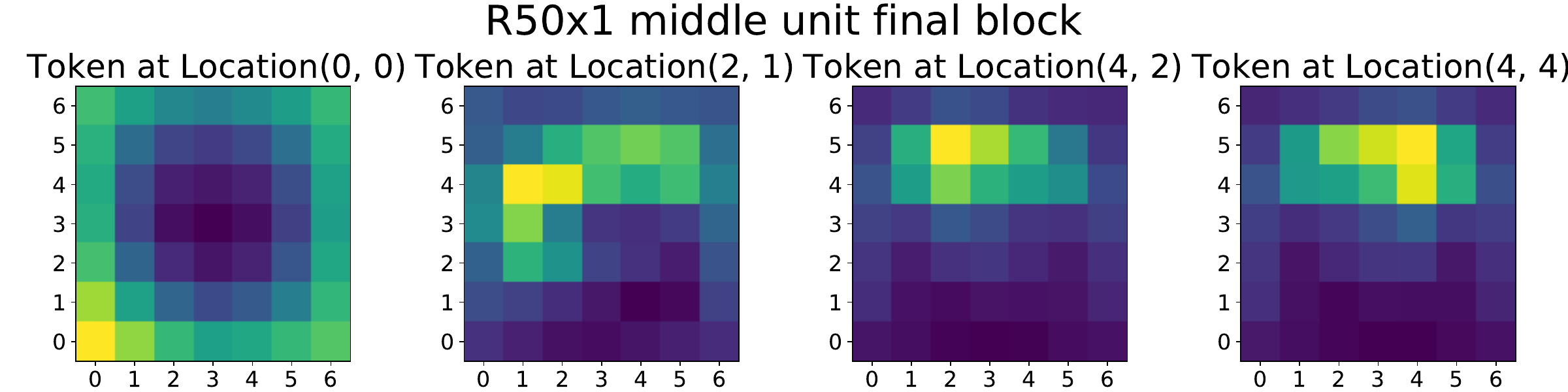} \\
    \includegraphics[width=\columnwidth]{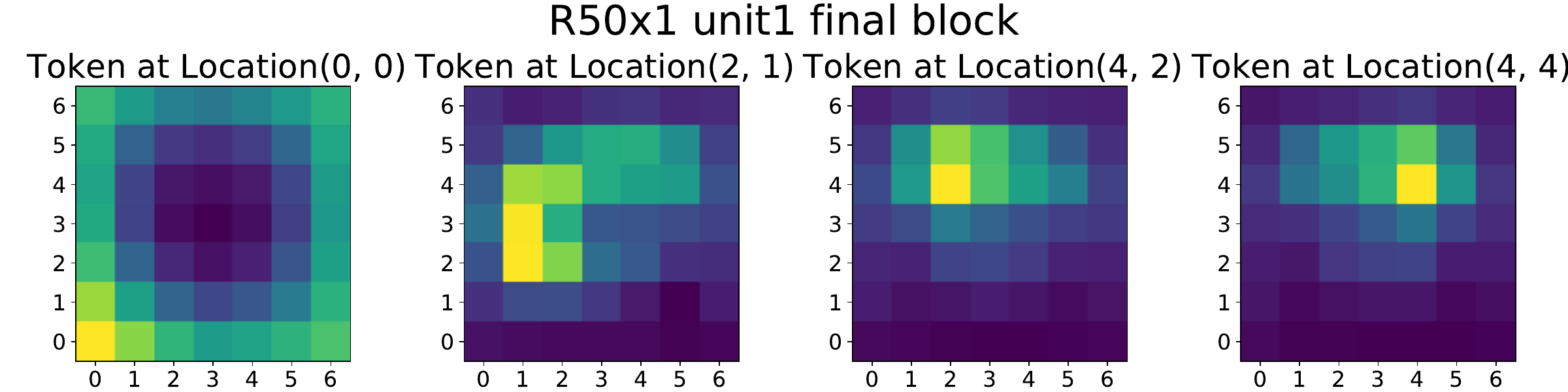}
    \caption{\small \textbf{Localization heatmaps for layers of ResNet below final layer}. Comparing to Figure \ref{fig:localization-heatmaps} in the main text, we see that layers in the ResNet lower than the final layer display better localization, but still not as clear as the CLS trained ViT models.}
    \label{fig:localization-heatmaps-resnet-app}
\end{figure}

\begin{figure}
    \centering
    \begin{tabular}{c} \includegraphics[height=0.2\columnwidth]{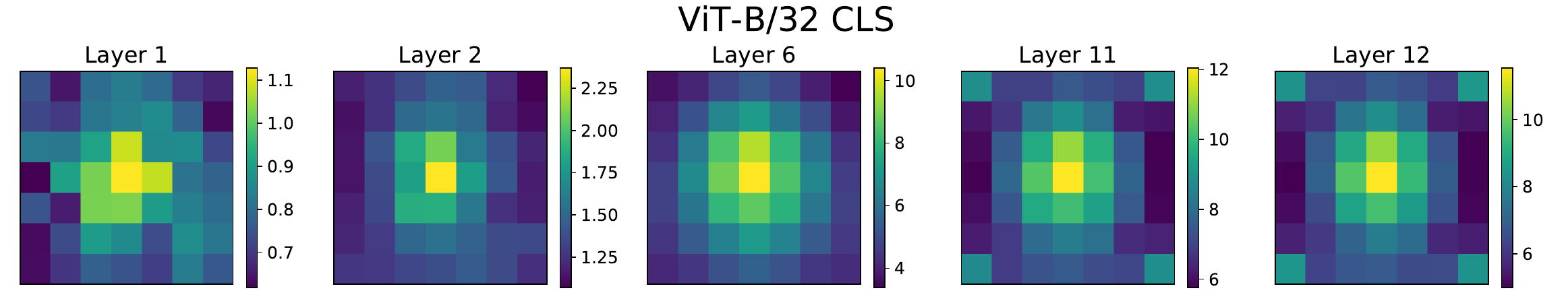} \\
    \includegraphics[height=0.2\columnwidth]{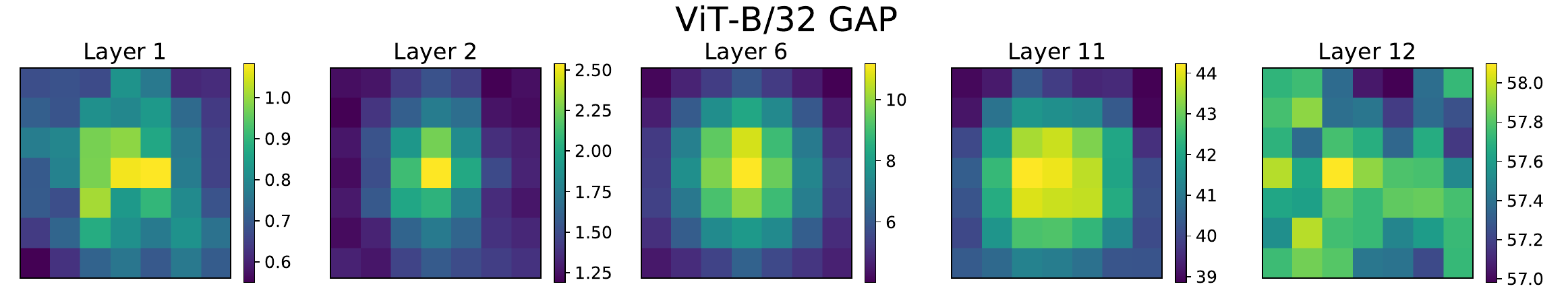} \\
    \includegraphics[height=0.2\columnwidth]{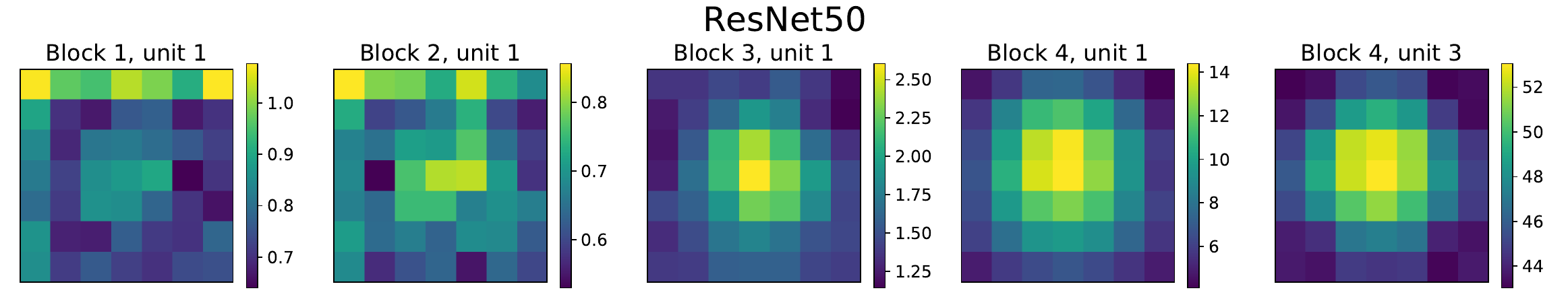} \\
    \end{tabular}
    \caption{Linear probes spatial localization. We train a linear probe on each individual token and plot the average accuracy over the test set, in percent. Here we plot the results for each token a subset of layers in 3 models: ViT-B/32 trained with a classification token (CLS) or global average pooling (GAP), as well as a ResNet50. Note the different scales of values in different sub-plots.}
    \label{fig:linear_probes-localization-spatial}
\end{figure}

\FloatBarrier

\section{Additional Representation Propagation Results}
\begin{figure}[h]
    \centering
    \includegraphics[width=\columnwidth]{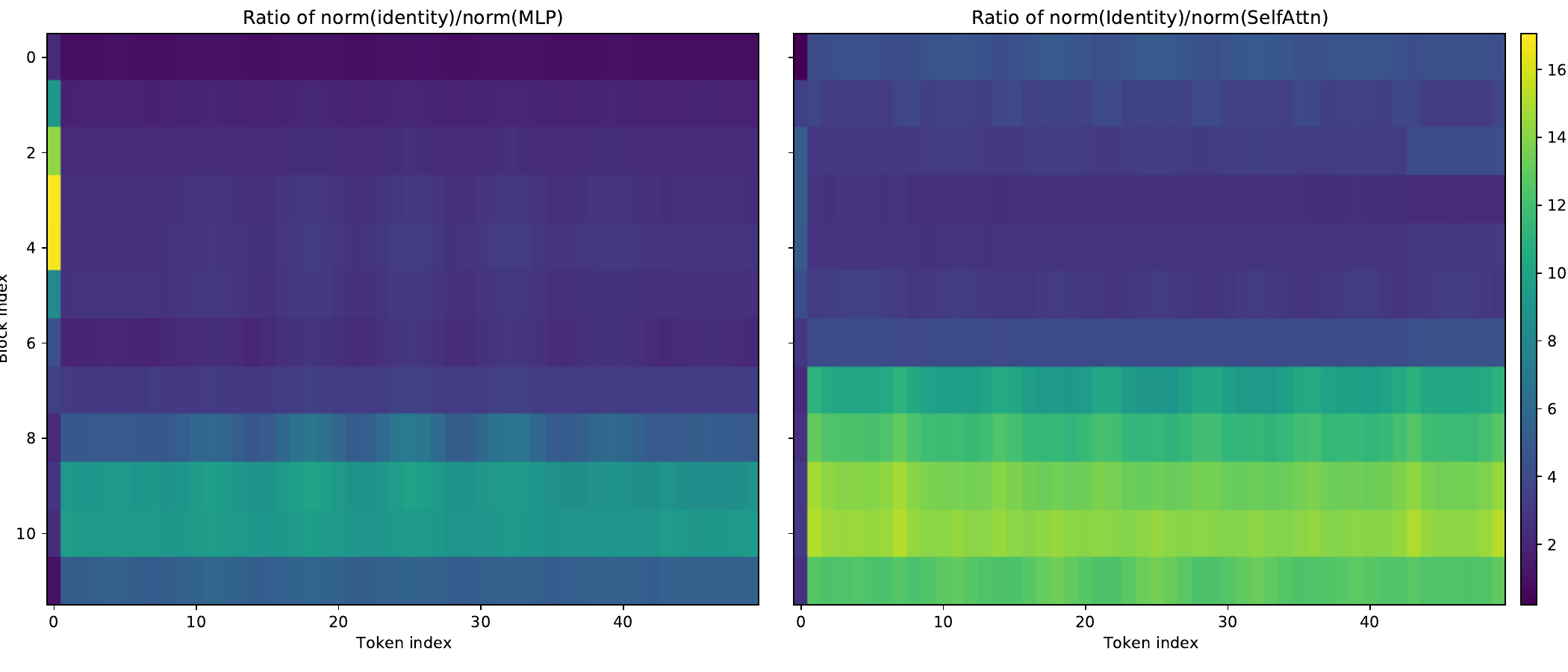}
    \caption{\small \textbf{Additional Heatmaps of Representation Norms} Ratio Representation Norms  $||z_i|| / || f(z_i)||$ between skip connection and the MLP or Self-Attention block for the hidden representation of each block on ViT-B/32, separately for each Token (Token 0 is CLS).}
    \label{fig:identity-connection-normplots-full}
\end{figure}

\begin{figure}
    \centering
    \begin{tabular}{cc}
     \includegraphics[width=1.0\columnwidth]{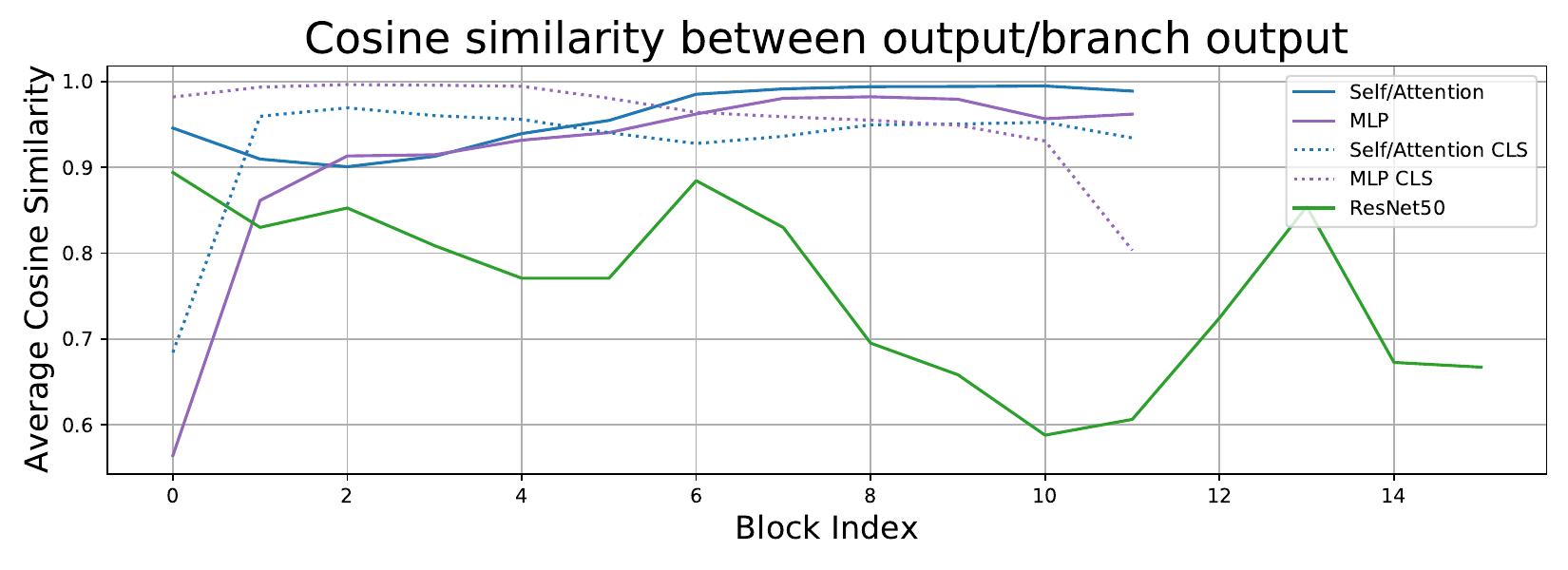}
    \end{tabular}
    \caption{\textbf{Most information in ViT passes through Identity Connections}. Cosine Similarity of representations between the skip-connection (identity) and the longer branch for ViT-B/16 trained on ImageNet and a ResNet v1. For ViT, we show the CLS token separately from the rest of the representation.}
    \label{fig:identity-connection-cosine-similarity-plot}
\end{figure}

Figure~\Ref{fig:identity-connection-normplots-full} shows the ratio of representation norms between skip connections and MLP and Self-Attention Blocks. In both cases, we observe that the CLS token representation is mostly unchanged in the first few layers, while later layers change it rapidly, mostly via MLP blocks. The reverse is true for the spatial tokens representing image patches, whose representation is mostly changed in earlier layers and does not change much during later layers. Looking at the cosine similarity of representations between output in Figure~\Ref{fig:identity-connection-cosine-similarity-plot} confirms these findings: while spatial token representations change more in early layers, the output of later blocks is very similar to the representation present on the skip connections, while the inverse is true for the CLS token.

\FloatBarrier

\section{Additional results on linear probes}
\label{sec:app-linear-probes}
\begin{figure}
    \centering
    \begin{tabular}{ccc}
    \hspace{-10mm} \includegraphics[height=0.3\columnwidth]{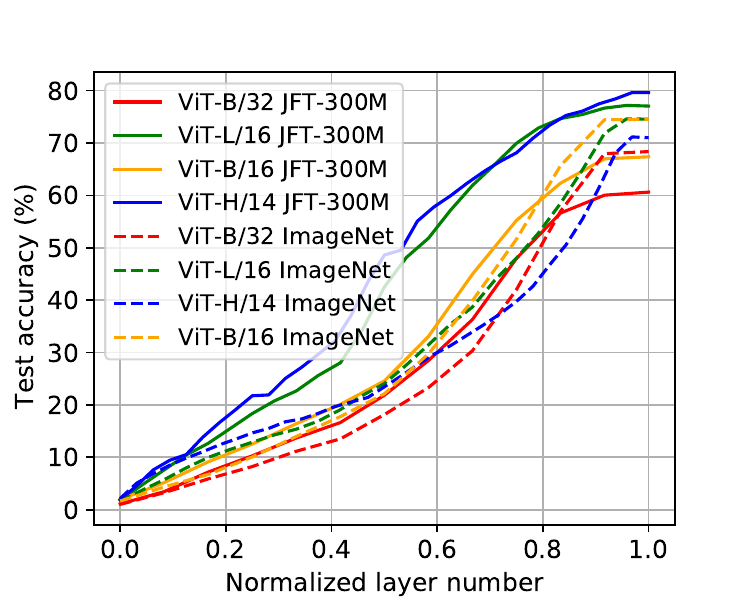} &
    \hspace{-7mm} \includegraphics[height=0.3\columnwidth]{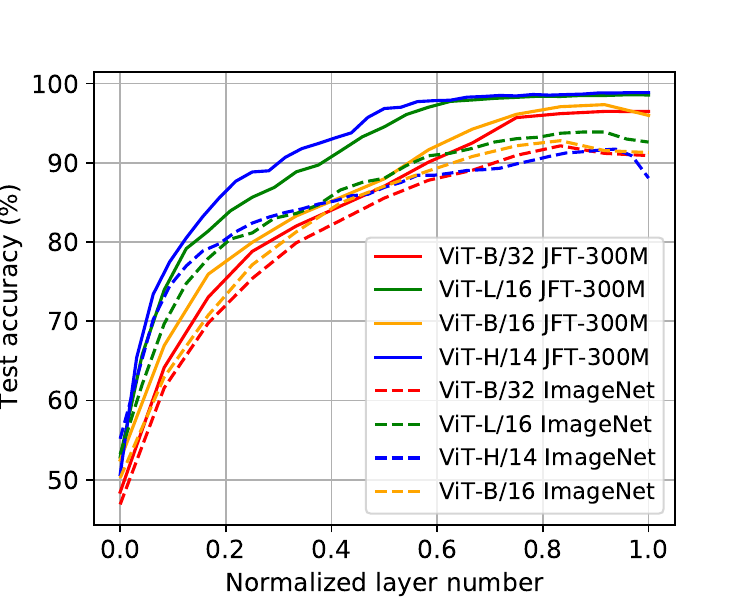} &
    \hspace{-7mm} \includegraphics[height=0.3\columnwidth]{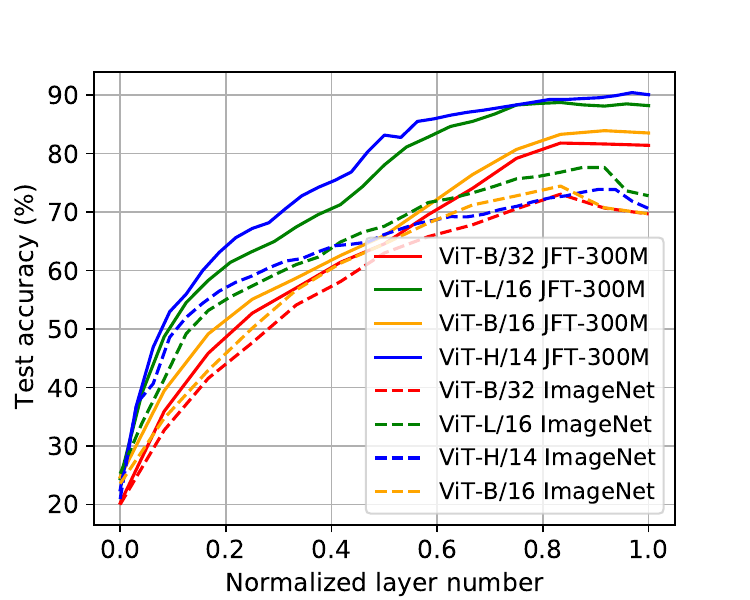} \\
    \hspace{-10mm} (a) ImageNet &
    \hspace{-7mm} (b) CIFAR-10 &
    \hspace{-7mm} (c) CIFAR-100 \\
    \end{tabular}
    \caption{Experiments with linear probes. Additional results on models pre-trained on JFT-300M and ImageNet (Fig.~\ref{fig:linear_probes-1} left)~-- with the addition of ViT-B models and CIFAR-10/100 datasets.}
    \label{fig:linear_probes-supp-jft-vs-imagenet}
\end{figure}

\begin{figure}
    \centering
    \begin{tabular}{cc}
    \includegraphics[height=0.4\columnwidth]{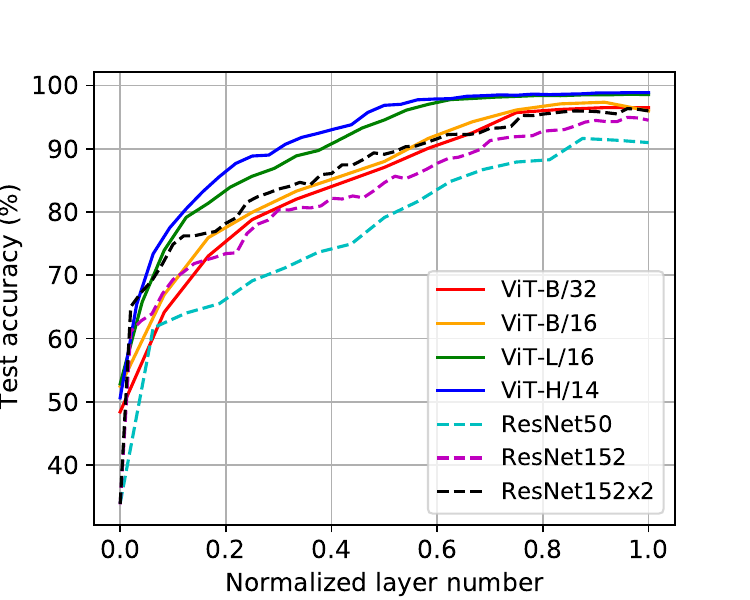} &
    \includegraphics[height=0.4\columnwidth]{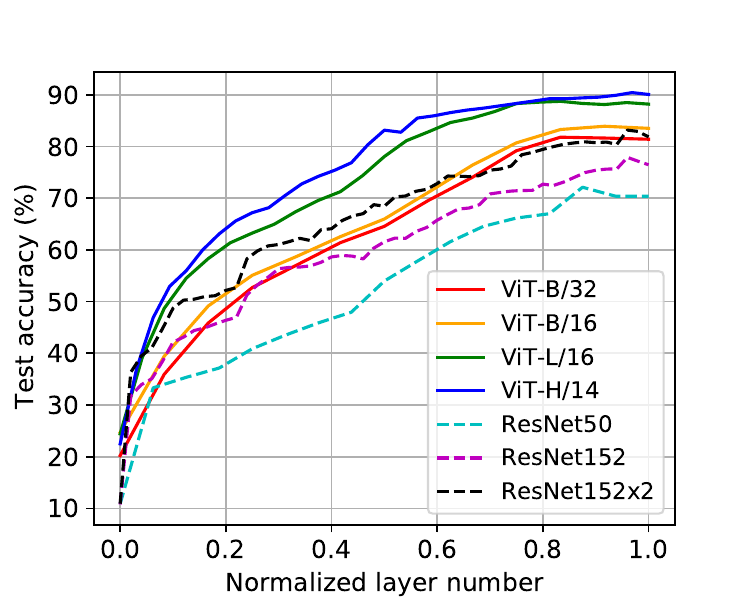} \\
    (a) CIFAR-10 &
    (b) CIFAR-100 \\
    \end{tabular}
    \caption{Experiments with linear probes. Additional results on comparison of ViT and ResNet models (Fig.~\ref{fig:linear_probes-1} right) on CIFAR-10/100 datasets.}
    \label{fig:linear_probes-supp-vit-vs-resnet}
\end{figure}

\begin{figure}
    \centering
    \begin{tabular}{cc}
    \includegraphics[height=0.4\columnwidth]{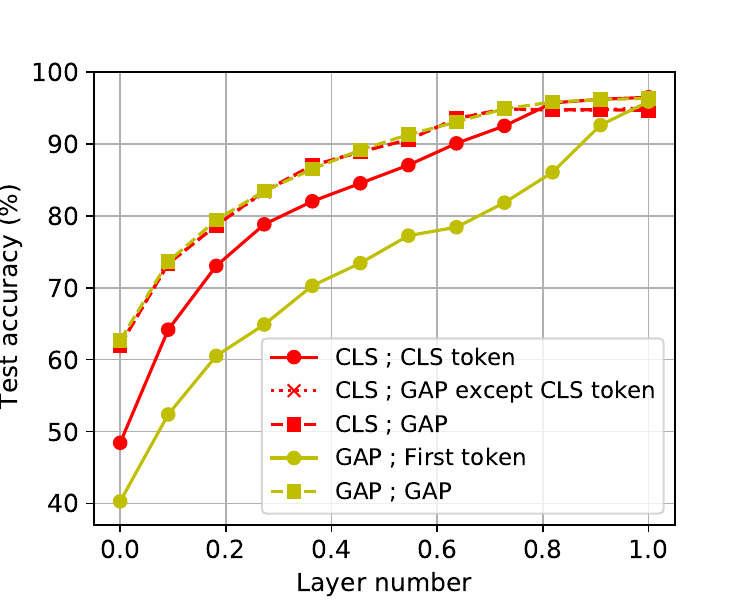} &
    \includegraphics[height=0.4\columnwidth]{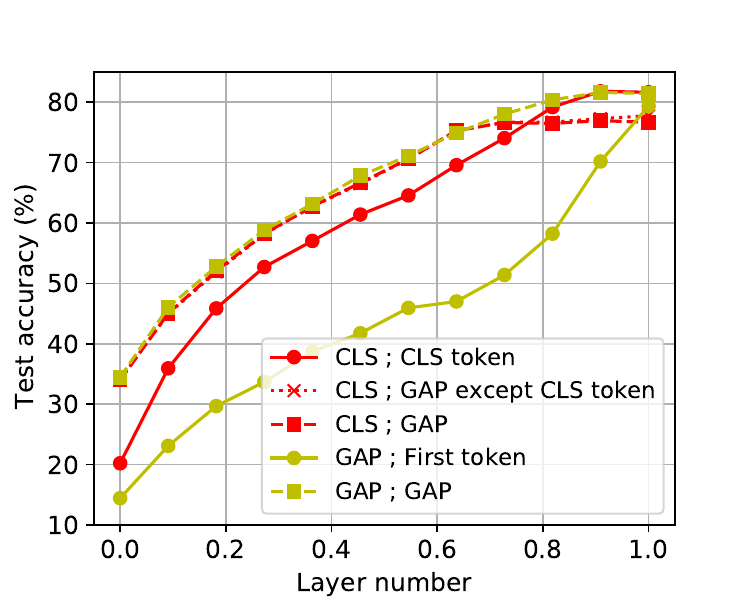} \\
    (a) CIFAR-10 &
    (b) CIFAR-100 \\
    \end{tabular}
    \caption{Experiments with linear probes. Additional results on comparison of ViT and ResNet models (Fig.~\ref{fig:linear_probes-2} right) on CIFAR-10/100 datasets.}
    \label{fig:linear_probes-supp-cls-vs-gap}
\end{figure}
Here we provide additional results on linear probes complementing Figures~\ref{fig:linear_probes-2} and~\ref{fig:linear_probes-1} of the main paper.
In particular, we repeat the linear probes on the CIFAR-10 and CIFAR-100 datasets and, in some cases, add more models to comparisons.
For CIFAR-10 and CIFAR-100, we use the first 45000 images of the training set for training and the last 5000 images from the training set for validation. 
Additional results are shown in Figures~\ref{fig:linear_probes-supp-jft-vs-imagenet}, ~\ref{fig:linear_probes-supp-vit-vs-resnet}, ~\ref{fig:linear_probes-supp-cls-vs-gap}.


Moreover, we discuss the results in the main paper in more detail.
In Figure~\ref{fig:linear_probes-2} (left) we experiment with different ways of evaluating a ViT-B/32 model. 
We vary two aspects: 1) the classifier with which the model was trained, classification token (CLS) or global average pooling (GAP), 2) The way the representation is aggregated: by just taking the first token (which for the CLS models is the CLS token), averaging all tokens, or averaging all tokens except for the first one.

There are three interesting observations to be made. 
First, CLS and GAP models evaluated with their ``native'' representation aggregation approach~-- first token for CLS and GAP for GAP~-- perform very similarly.
Second, the CLS model evaluated with the pooled representation performs on par with the first token evaluation up to last several layers, at which point the performance plateaus.
This suggests that the CLS token is crucially contributing to information aggregation in the latter layers.
Third, linear probes trained on the first token of a model trained with a GAP classifier perform very poorly for the earlier layers, but substantially improve in the latter layers and almost match the performance of the standard GAP evaluation in the last layer.
This suggests all tokens are largely interchangeable in the latter layers of the GAP model.

To better understand the information contained in individual tokens, we trained linear probes on all individual tokens of three models: ViT-B/32 trained with CLS or GAP, as well as ResNet50.
Figure~\ref{fig:linear_probes-2} (right) plots average performance of these per-token classifiers.
There are two main observations to be made.
First, in the ViT-CLS model probes trained on individual tokens perform very poorly, confirming that the CLS token plays a crucial role in aggregating global class-relevant information.
Second, in ResNet the probes perform poorly in the early layers, but get much better towards the end of the model. 
This behavior is qualitatively similar to the ViT-GAP model, which is perhaps to be expected, since ResNet is also trained with a GAP classifier.

\FloatBarrier

\section{Effects of Scale on Transfer Learning}
\label{sec:app-scale-transfer-learning}
\begin{figure}
    \centering
    \includegraphics[width=0.8\columnwidth]{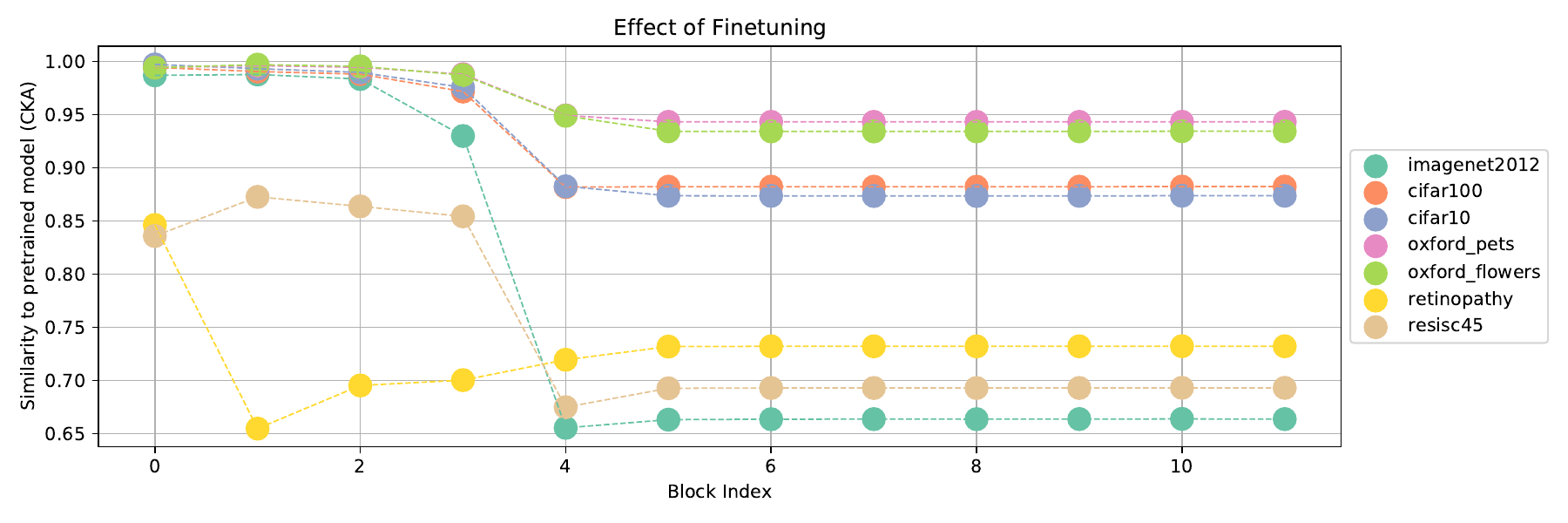}  \\
    \includegraphics[width=0.8\columnwidth]{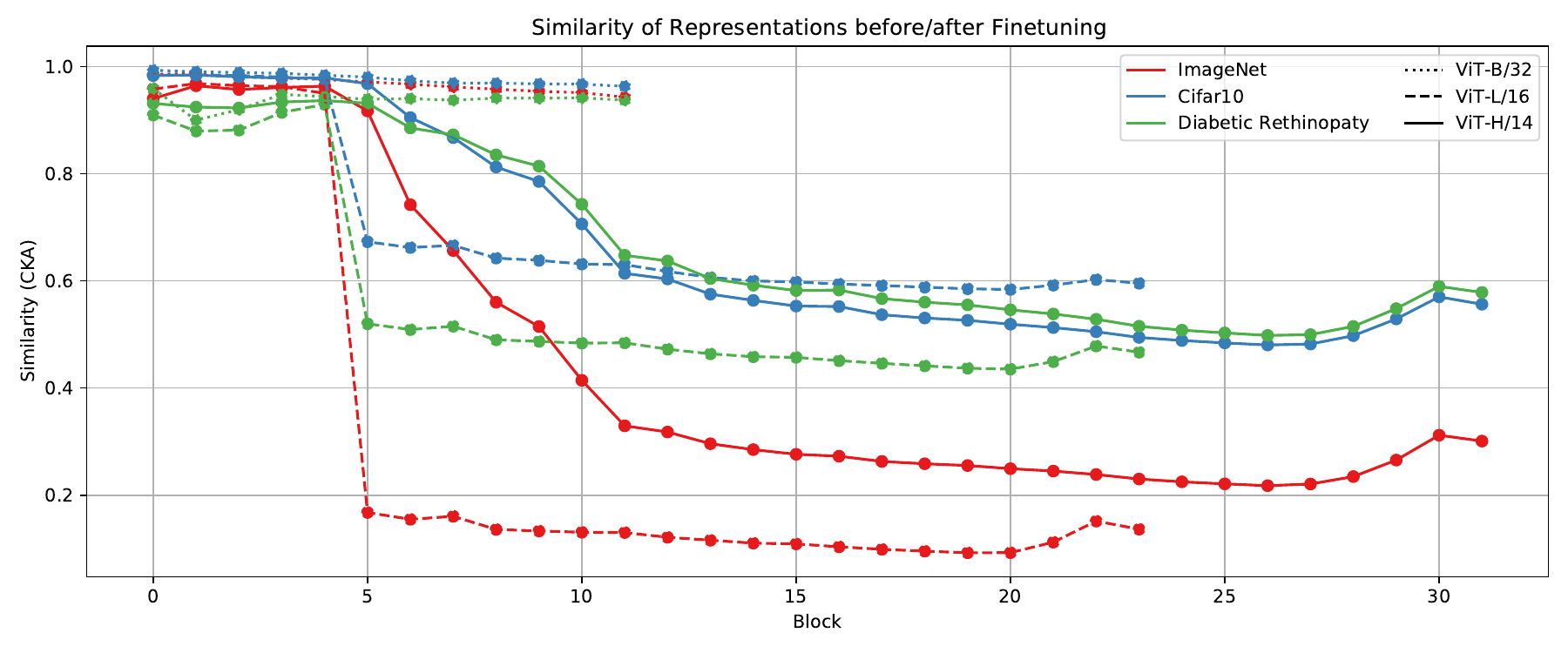}
    \caption{(top) Similarity of representations at each block for ViT-B/16 models compared to before finetuning. (bottom) Similarity of representations at each block for different ViT model sizes.}
    \label{fig:training-set-size-full}
\end{figure}

\begin{figure}
    \centering
    \includegraphics[width=0.8\columnwidth]{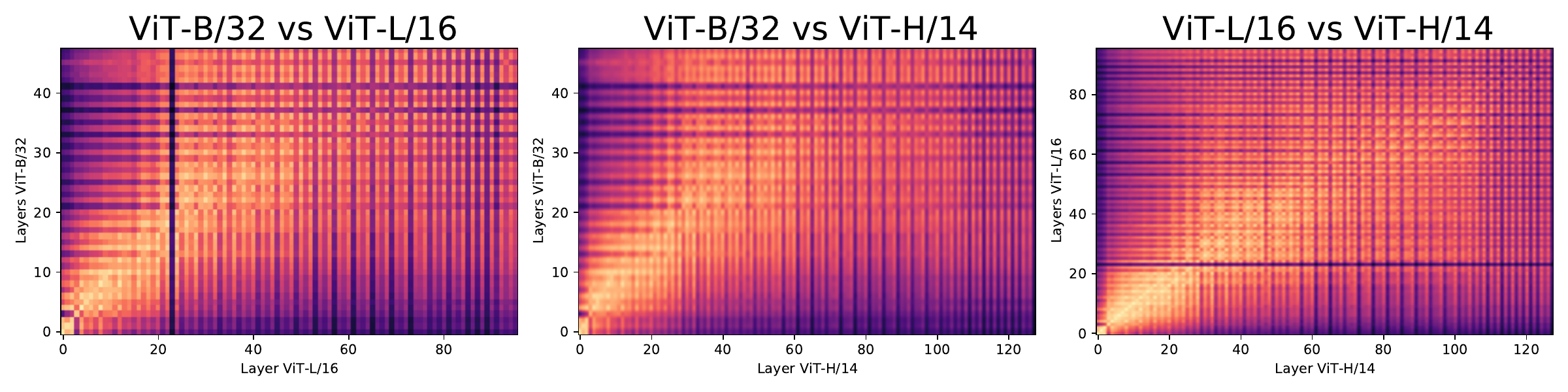} 
    \caption{Similarity of representations of different ViT model sizes.}
    \label{fig:vit-cross-model-cka}
\end{figure}

Finally, we study how much representations change through the finetuning process for a model pretrained on JFT-300M, finding significant variation depending on the dataset. For tasks like ImageNet or Cifar100 which are very similar to the natural images setting of JFT300M, the representation does not change too much. For medical data (Diabethic Retinopathy detection) or satellite data (RESISC45), the changes are more pronounced. In all cases, it seems like the first four to five layers remain very well preserved, even accross model sizes. This indicates that the features learned there are likely to be fairly task agnostic, as seen in Figure~\Ref{fig:training-set-size-full} and Figure~\Ref{fig:vit-cross-model-cka}.

\label{sec:transfer-learning-app}

\FloatBarrier
\section{Preliminary Results on MLP-Mixer}
\label{sec:mlp-mixer}
\begin{figure}[h!]
    \centering
    \includegraphics[width=0.8\columnwidth]{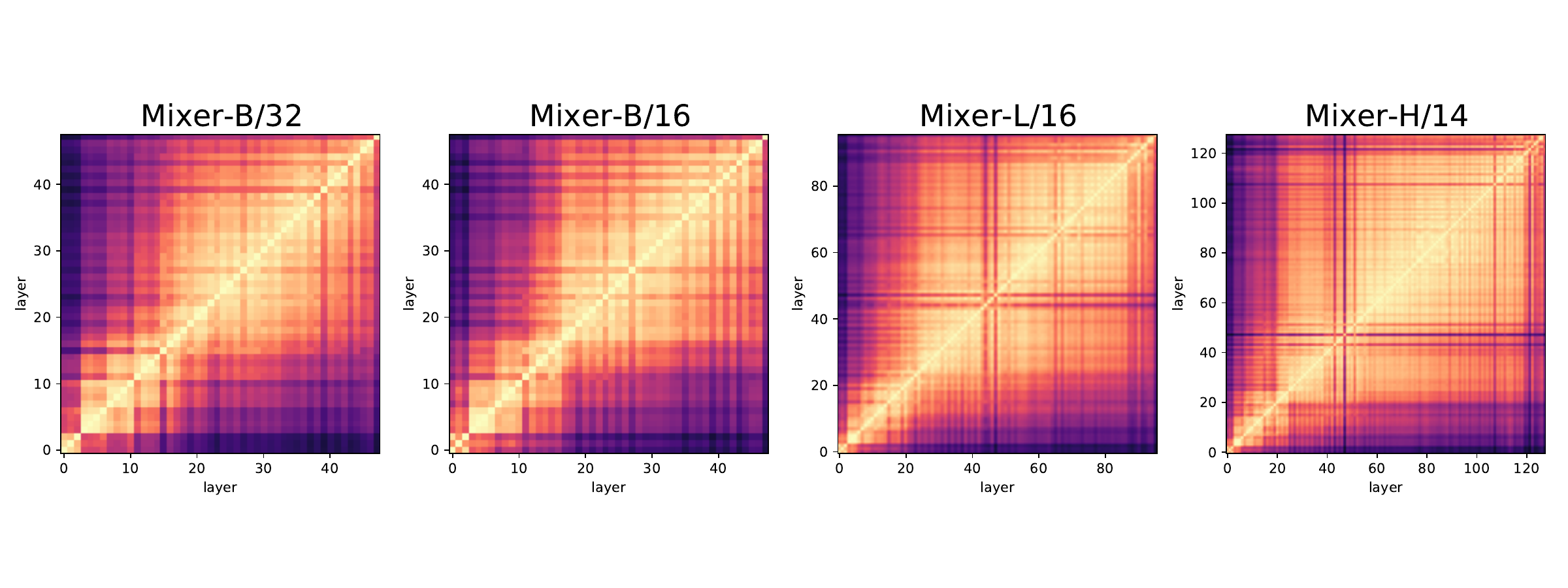} 
    \caption{Similarity of representations of different ViT model sizes.}
    \label{fig:mixer-models}
\end{figure}

\begin{figure}[htb]
    \centering
    \includegraphics[width=0.8\columnwidth]{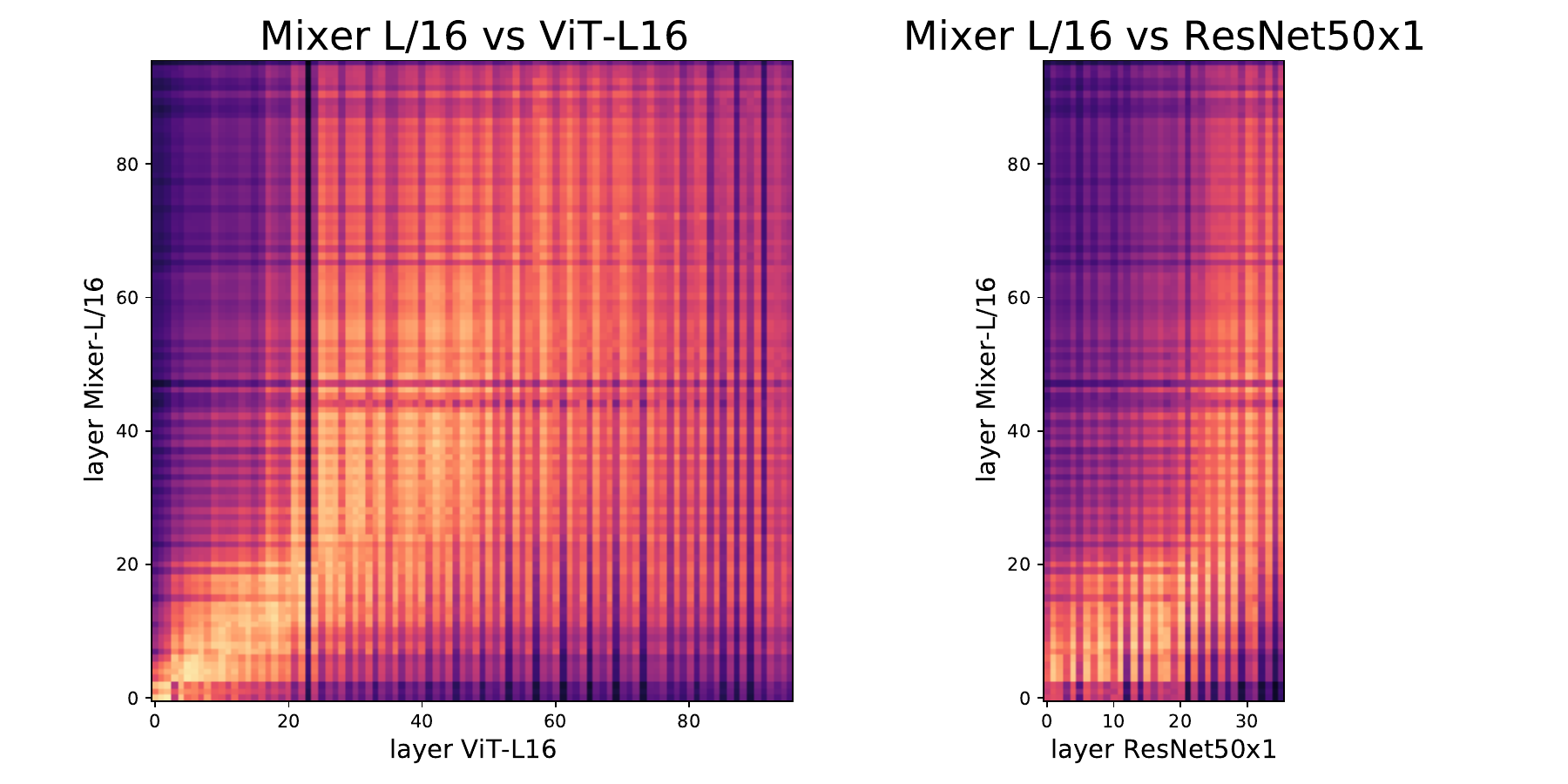} 
    \caption{Similarity of representations of different ViT model sizes.}
    \label{fig:mixer-comparison}
\end{figure}

Figure~\Ref{fig:mixer-models} shows the representations from various MLP-Mixer models. The representations seem to also fall very clearly into distinct, uncorrelated blocks, with a smaller block in the beginning and a larger block afterwards. This is independent of model size. Comparing these models with ViT or ResNet as in Figure~\Ref{fig:mixer-comparison} models makes it clear that overall, the models behave more similar to ViT than ResNets (c.f. Fig.~\Ref{fig:representation-structure} and \Ref{fig:representation-cross-comparisons}).


\end{document}